\documentclass{article}

\usepackage{microtype}
\usepackage{graphicx}
\usepackage{subfigure}
\usepackage{booktabs} 
\usepackage{multirow}
\usepackage{tikz}
\usepackage{threeparttable}
\usepackage{tikz-cd,mathtools}
\usepackage{mathtools}
\usepackage{floatpag}
\usepackage{float}

\usepackage{hyperref}

\usepackage[accepted]{icml2024}

\usepackage{amsmath}
\usepackage{amssymb}
\usepackage{mathtools}
\usepackage{amsthm}

\usepackage{url}
\usepackage{multirow}
\usepackage{subfigure}
\usepackage{graphicx}
\usepackage{wrapfig}
\usepackage{color}
\usepackage[T1]{fontenc}
\usepackage{marvosym}
\usepackage{bbm}

\theoremstyle{plain}

\theoremstyle{definition}

\theoremstyle{remark}

\usepackage[textsize=tiny]{todonotes}

\definecolor{MyDarkRed}{rgb}{0.8,0.02,0.02}
\definecolor{TODO}{rgb}{0.8,0,0}
\definecolor{revise}{rgb}{0,0,0}

\icmltitlerunning{HelmFluid: Learning Helmholtz Dynamics for Interpretable Fluid Prediction}

\begin{document}

\twocolumn[
\icmltitle{HelmFluid: Learning Helmholtz Dynamics for Interpretable Fluid Prediction}



\icmlsetsymbol{equal}{*}

\begin{icmlauthorlist}
\icmlauthor{Lanxiang Xing}{equal,tsinghua}
\icmlauthor{Haixu Wu}{equal,tsinghua}
\icmlauthor{Yuezhou Ma}{tsinghua}
\icmlauthor{Jianmin Wang}{tsinghua}
\icmlauthor{Mingsheng Long}{tsinghua}
\end{icmlauthorlist}

\icmlaffiliation{tsinghua}{School of Software, BNRist, Tsinghua University. Lanxiang Xing $<$xlx22@mails.tsinghua.edu.cn$>$}

\icmlcorrespondingauthor{Mingsheng Long}{mingsheng@tsinghua.edu.cn}

\icmlkeywords{Fluid Prediction, Operator Learning, Deep Learning}

\vskip 0.3in
]



\printAffiliationsAndNotice{\icmlEqualContribution} 
\begin{abstract}
Fluid prediction is a long-standing challenge due to the intrinsic high-dimensional non-linear dynamics. Previous methods usually utilize the non-linear modeling capability of deep models to directly estimate velocity fields for future prediction. However, skipping over inherent physical properties but directly learning superficial velocity fields will overwhelm the model from generating precise or physics-reliable results. In this paper, we propose the \emph{HelmFluid} toward an accurate and interpretable predictor for fluid. Inspired by the Helmholtz theorem, we design a \emph{HelmDynamics} block to learn \emph{Helmholtz dynamics}, which decomposes fluid dynamics into more solvable curl-free and divergence-free parts, physically corresponding to potential and stream functions of fluid. By embedding the HelmDynamics block into a \emph{Multiscale Multihead Integral Architecture}, HelmFluid can integrate learned Helmholtz dynamics along temporal dimension in multiple spatial scales to yield future fluid. Compared with previous velocity estimating methods, HelmFluid is faithfully derived from Helmholtz theorem and ravels out complex fluid dynamics with physically interpretable evidence. Experimentally, HelmFluid achieves consistent state-of-the-art in both numerical simulated and real-world observed benchmarks, even for scenarios with complex boundaries. Code is available at \href{https://github.com/thuml/HelmFluid}{https://github.com/thuml/HelmFluid}.
\end{abstract}

\section{Introduction}

Fluid is one of the basic substances in the physical world. Its prediction is of immense importance in extensive real-world applications, such as atmospheric prediction for weather forecasting and airflow modeling for airfoil design, which has attracted significant attention from both science and engineering areas. However, it is quite challenging to capture and predict the intricate high-dimensional non-linear dynamics within the fluid due to imperfect observations, coupled multiscale interactions, etc. In this paper, we focus on a more practical scenario that predicts future states of fluid from \emph{partially observed} physical quantities.

\begin{figure*}[t]
\begin{center}
\centerline{\includegraphics[width=1.0\textwidth]{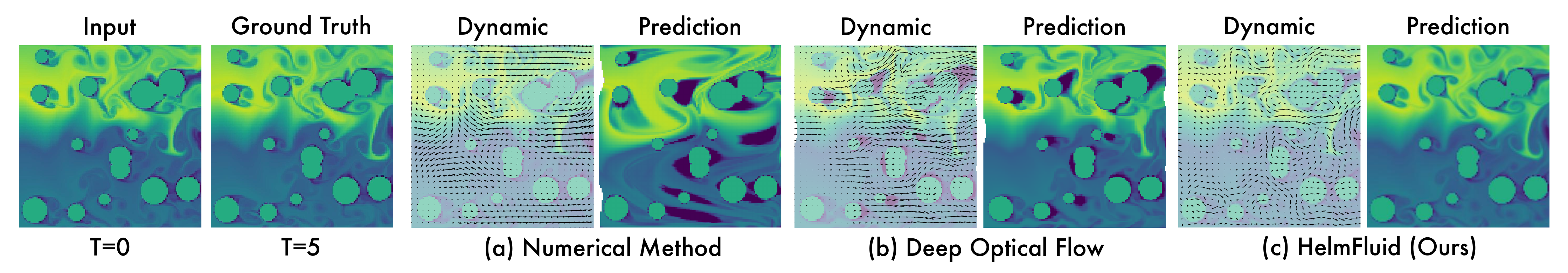}}
\vspace{-5pt}
    \caption{Comparison on dynamics and fluid modeling. Different from the numerical method \citep{ruzanski2011casa} and optical-flow-based deep model \citep{sun2018pwc}, HelmFluid infers the dynamics from the inherent physics quantities: potential and stream functions.}
    \label{fig:introduction}
\end{center}
\vspace{-20pt}
\end{figure*}

Recently, deep models have achieved impressive progress in solving complex physical systems~\citep{karniadakis2021physics,wang2023scientific}. One paradigm is learning neural operators to directly predict the future fluid field based on past observations \citep{lu2021learning,li2021fourier,wu2023solving}. These methods focus on leveraging the non-linear modeling capacity of deep models to approximate complex mappings between past and future fluids. However, directly learning neural operators may fail to generate interpretable evidence for prediction results and incur uncontrolled errors. Another mainstreaming paradigm attempts to estimate the dynamic fields of fluid with deep models for future prediction. It is notable that the superficial dynamics are actually driven by underlying physical rules. Directly estimating the velocity fields regarding less physical properties may overwhelm the model from generating precise and plausible prediction results~\citep{sun2018pwc,zhang2022learning}. As shown in Figure \ref{fig:introduction}, it is hard to directly capture the complex dynamics of fluid, where the learned dynamics will be too tanglesome to guide the fluid prediction.

To tackle the above challenges, we attempt to capture the intricate dynamics with physical insights for accurate and interpretable fluid prediction. In this paper, we dive into the physical properties of fluid and propose the \emph{Helmholtz dynamics} as a new paradigm to represent fluid dynamics. Concretely, Helmholtz dynamics is inspired by the Helmholtz theorem \citep{van1958helmholtz} and attributes the intricate dynamics to the potential and stream functions of fluid, which are intrinsic physical quantities of fluid and can directly derive the curl-free and divergence-free parts of fluid respectively. Compared with superficial velocity fields, our proposed Helmholtz dynamics decompose the intricate dynamics into more solvable components, thereby easing the dynamics learning process of deep models. Besides, this new dynamics requires the model to learn the inherent properties of fluid explicitly, which also empowers the prediction with endogenetic physical interpretability. 

Based on the above ideas, we present the \emph{HelmFluid} model with novel \emph{HelmDynamics} blocks to capture the Helmholtz dynamics for interpretable fluid prediction. HelmDynamics is faithfully implemented from the Helmholtz decomposition, which can separately estimate the potential and stream functions of fluid from learned spatiotemporal correlations and further derive curl-free and divergence-free velocities. As a flexible module, HelmDynamics can conveniently encode boundary conditions into the correlation calculation process and adapt to complex boundary settings in multifarious real-world applications. Further, we design the \emph{Multiscale Multihead Integral Architecture} in HelmFluid to fit the multiscale nature of fluid, which can integrate Helmholtz dynamics learned by HelmDynamics blocks along temporal dimension in multiple spatial scales to predict the future fluid. Experimentally, HelmFluid achieves consistent state-of-the-art in various scenarios, covering both synthetic and real-world benchmarks with complex boundary settings. Our contributions are summarized in the following:
\vspace{-5pt}
\begin{itemize}
    \item Inspired by the Helmholtz theorem, we propose the \emph{Helmholtz dynamics} to attribute intricate dynamics into inherent properties of fluid, which decomposes intricate dynamics into more solvable parts and empowers the prediction process with physical interpretability.
    \item We propose \emph{HelmFluid} with the \emph{HelmDynamics block} to capture Helmholtz dynamics. By integrating learned dynamics along temporal dimension with the \emph{Multiscale Multihead Integral Architecture}, HelmFluid can predict future fluid with physically plausible evidence.
    \item HelmFluid achieves consistent state-of-the-art in extensive benchmarks, covering both synthetic and real-world datasets, as well as various boundary conditions.
\end{itemize}

\vspace{-5pt}
\section{Related Work}

As a foundation problem in science and engineering areas, fluid prediction has been widely explored. Traditional methods can solve Navier-Stokes equations with numerical algorithms, while they may fail in the real-world fluid due to imperfect observations of initial conditions and inaccurate estimation of equation parameters. Besides, these numerical methods also suffer from huge computation cost. Recently, owing to the great non-linear modeling capacity, data-driven deep models for fluid prediction have attached substantial interests, which can be roughly categorized into the following paradigms according to whether learning velocity fields explicitly or not.

\vspace{-5pt}
\paragraph{Neural fluid simulator} This paradigm of works attempts to directly generate future fluid with deep models. One direction is formulating partial differential equations (PDEs) along with initial and boundary conditions as loss function terms, and parameterizing the solution as a deep model \citep{evans2010partial,raissi2019physics,raissi2020hidden,lu2021deepxde}. These approaches rely highly on exact physics equations, thereby suffering from imperfect observations and inherent randomness in real-world applications. Another branch of methods does not require the exact formulation of governing PDEs. They attempt to learn neural operators to approximate complex input-output mappings in scientific tasks, which enables the prediction of future fluid solely based on past observations. For example, \cite{lu2021learning} proposed DeepONet in a branch-trunk framework with proven universal approximation capability. FNO \citep{li2021fourier} approximates the integral operator through a linear transformation in the Fourier domain. Afterward, U-NO \citep{rahman2023uno} enhances FNO with a multi-scale framework. Later, \citet{wu2023solving} proposed latent spectral models (LSM) to solve high-dimensional PDEs in the latent space by learning multiple basis operators. Still, these methods may fail to provide interpretable evidence for prediction results, such as intuitive physics quantities or visible velocity fields. Going beyond the above-mentioned methods, we propose HelmFluid as a purely data-driven model but with special designs to enhance physical interpretability.

\vspace{-5pt}
\paragraph{Fluid dynamics modeling} Estimating velocity fields is a direct and intuitive way of predicting the future fluid. Typically, optical flow \citep{horn1981determining} is proposed to describe the motion between two successive observations.
Recently, many deep models have been proposed to estimate optical flow, such as PWC-Net \citep{sun2018pwc} and RAFT \citep{teed2020raft}. However, since the optical flow was originally designed for rigid bodies, it struggles seriously in capturing fluid motion and will bring serious accumulation errors in the prediction process. Especially for fluid, \citeauthor{zhang2022learning}~incorporated physical constraints from Navier-Stokes equations to refine the velocity field predicted by PWC-Net \citep{sun2018pwc} and further embedded the advection-diffusion equation into the deep model to predict the future fluid. Recently, Vortex \cite{deng2023learning} ensembles the observable Eulerian flow and the hidden Lagrangian vortical evolution to capture the intricate dynamics within the fluid. Unlike previous works, we propose to learn the inherent physical quantities of Helmholtz dynamics that explicitly derive the velocity fields, and further predict the future fluid with the Runge--Kutta temporal integral. This decomposes the intricate dynamics into more solvable components and facilitates our model with physical interpretability.

\vspace{-5pt}
\paragraph{Computer graphics for fluid simulation} Solving Navier-Stokes equations with learning-based computer graphics methods often uses a stream function paradigm to enforce the incompressibility condition \citep{ando2015stream}. \citeauthor{kim2019deep} successfully synthesized plausible and divergence-free 2D and 3D fluid velocities from a set of reduced parameters but requiring ground truth velocity supervision, a rarity in real-world data. Recently, \citeauthor{liu2021discovering} estimated the underlying physics of advection-diffusion equations, incorporating ground truth velocity and diffusion tensors supervision. \citeauthor{franz2023learning} simulated a realistic 3D density and velocity sequence from single-view sequences without 3D supervision, but it is not designed for predictive tasks as it utilizes future information to calculate current density. Unlike previous methods, our method learns the velocity {field} end-to-end from {physical quantities observed in the past via} Helmholtz dynamics, {relying neither on} ground truth velocity supervision {nor on} stream function. Such an unsupervised paradigm enables our model to capture more intricate fluid dynamics and extends its capability to a broader range of scenarios.

\section{HelmFluid}

In this paper, we highlight the key components of fluid prediction as providing physical interpretability and handling intricate dynamics. To achieve these objectives, we present the HelmFluid model with \emph{HelmDynamics} blocks to capture the \emph{Helmholtz dynamics} for 2D fluid, which is inspired by the Helmholtz theorem and attributes superficial complex dynamics to the inherent properties of fluid. Further, we design the \emph{Multiscale Multihead Integral Architecture} to integrate the learned dynamics along the temporal dimension in multiple scales to predict the future states of fluid.

\subsection{Learning Helmholtz Dynamics}

Learning intricate velocity fluid directly from data may overwhelm the model. Hence, we propose to learn Helmholtz dynamics via a HelmDynamics block, which is a faithful implementation of the the Helmholtz theorem that decomposes complex fluid dynamics into more solvable components.

\begin{figure*}[t]
\begin{center}
\centerline{\includegraphics[width=\textwidth]{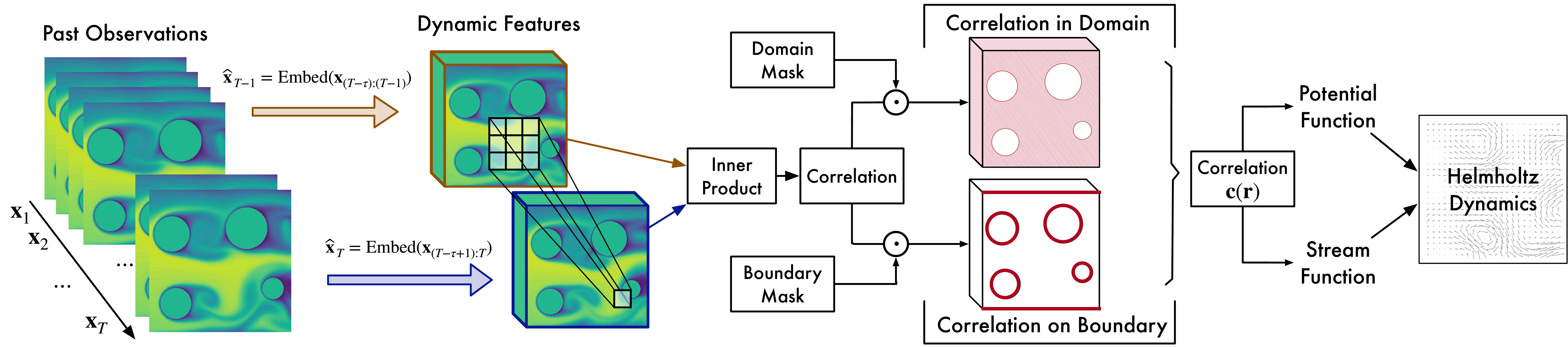}}
\vspace{-5pt}
    \caption{HelmDynamics block, which learns spatiotemporal correlations $\mathbf{c}(\mathbf{r})$ both in the domain and on the boundary to estimate potential and stream functions of fluid from past observations for composing the Helmholtz dynamics.}
    \label{fig:helmdynamics}
    \vspace{-20pt}
\end{center}
\end{figure*}

\paragraph{Helmholtz decomposition theorem}
Helmholtz decomposition \citep{van1958helmholtz} plays an important role in simulating fluid dynamics, which can decompose a dynamic field into a curl-free component and a divergence-free component for simplification, and is highly related to the solvability theory of Navier-Stokes equations \citep{morrison2013introduction}. 

Given a 3D dynamic field ${\mathbf{F}}:\mathbb{V}\rightarrow \mathbb{R}^3$ with a bounded domain $\mathbb{V}\subseteq\mathbb{R}^3$, we can obtain the following decomposition based on the Helmholtz theorem:
\begin{equation}
    \label{equ:helmholtz1}
    \begin{aligned}
    {\mathbf{F}}(\mathbf{r}) &= \nabla\Phi(\mathbf{r}) + \nabla \times {\mathbf{A}}(\mathbf{r}), \; \mathbf{r} \in \mathbb{V}.
    \end{aligned}
\end{equation}
It is notable that $\Phi: \mathbb{V}\rightarrow \mathbb{R}$ denotes the \emph{potential function}, which is a scalar field with its gradient field $\nabla\Phi$ representing the curl-free part of ${\mathbf{F}}$ guaranteed by $\nabla\times(\nabla\Phi)=\textbf{0}$. And ${\mathbf{A}}: \mathbb{V}\rightarrow \mathbb{R}^3$ denotes the \emph{stream function}, which is a vector field with $\nabla \times {\mathbf{A}}$ represents the divergence-free part of ${\mathbf{F}}$ guaranteed by $\nabla(\nabla\times\mathbf{A})=\textbf{0}$, thereby also indicating the incompressibility of the flow field.

\paragraph{Helmholtz dynamics for 2D fluid} 
Following mainstream works and conventional settings \citep{li2021fourier}, and for conciseness of presentation, this paper presents the \emph{HelmFluid} model on 2D fluid prediction. We project the Helmholtz theorem into 2D space by restricting the $z$-axis component of $\mathbf{F}$ to $0$, i.e.~$\mathbf{F}(\mathbf{r}) = \left(\mathbf{F}_x(\mathbf{r}\right), \mathbf{F}_y(\mathbf{r}), 0)^{\sf T}$. This restriction also vanishes the components of the stream function along $x$-axis and $y$-axis, namely $\mathbf{A}(\mathbf{r}) = \left((0, 0, \mathbf{A}_z(\mathbf{r})\right)^{\sf T}$, indicating that the stream function degenerates to a scalar field. Generalizations and experiments on more practical 3D fluid prediction are included in Appendix \ref{sec: 3DHelmFluid}.

According to the Helmholtz decomposition theorem (Eq.~\ref{equ:helmholtz1}), the fluid dynamics can be equivalently decomposed into \emph{curl-free} and \emph{divergence-free} parts for simplification. Thus, we define Helmholtz dynamics ${\mathbf{F}}_{\text{Helm}}(\Phi, \mathbf{A})$ explicitly as the function of potential and stream functions, which are inherent physics quantities of fluid. Concretely, for a 2D fluid defined in the domain $\mathbb{V}\subseteq\mathbb{R}^2$, its Helmholtz dynamics ${\mathbf{F}}_{\text{Helm}}$ can be formalized by potential function $\Phi: \mathbb{V}\to\mathbb{R}$ and stream function $\mathbf{A}: \mathbb{V}\to\mathbb{R}$ of fluid as follows:
\begin{equation}
    \label{helmholtz2}
    \begin{aligned}
    &{\mathbf{F}}_{\text{Helm}}(\Phi, \mathbf{A}) = \nabla\Phi + \nabla \times {\mathbf{A}} \\
    =& \underbrace{\left(\frac{\partial \Phi}{\partial x}, \frac{\partial \Phi}{\partial y}\right)}_{\text{Curl-free Velocity}} + \underbrace{\left(\frac{\partial {\mathbf{A}}}{\partial y}, -\frac{\partial {\mathbf{A}}}{\partial x}\right)}_{\text{Divergence-free Velocity}}.
    \end{aligned}
\end{equation}
According to the Helmholtz theorem (Eq.~\ref{equ:helmholtz1}), the function value of ${\mathbf{F}}_{\text{Helm}}$ is equivalent to the real dynamic field $\mathbf{F}$ but is more tractable.
By incorporating $\Phi$ and $\mathbf{A}$, Helmholtz dynamics naturally decomposes the intricate fluid into more solvable components and ravels out the complex dynamics into intrinsic physics quantities, thus benefiting more interpretable dynamics modeling~\citep{Bhatia2013}.

\paragraph{HelmDynamics block} To learn the Helmholtz dynamics, we propose the HelmDynamics block to estimate the potential and stream functions from past observations. As shown in Figure \ref{fig:helmdynamics}, we first embed input observations into two successive deep representations to keep the temporal dynamics information explicitly. Given a sequence of $T$ frames $\mathbf{x}=[\mathbf{x}_{1}, \cdots, \mathbf{x}_{T}], \mathbf{x}_i\in\mathbb{R}^{H\times W}$ successively observed from 2D fluid, this process can be written as
\begin{equation}
    \label{equ:embed}
    \begin{aligned}
    \widehat{\mathbf{x}}_{T-1}&=\operatorname{Embed}\left(\mathbf{x}_{(T-\tau):(T-1)}\right)\\
    \widehat{\mathbf{x}}_{T}&=\operatorname{Embed}\left(\mathbf{x}_{(T-\tau+1):T}\right),
    \end{aligned}
\end{equation}
where $\widehat{\mathbf{x}}_{T-1}, \widehat{\mathbf{x}}_{T}\in\mathbb{R}^{d_{\text{model}}\times H\times W}$ are the feature tensors at timestamps $T-1$ and $T$ respectively. Here, we embed the observations from a $\tau$ lookback window to capture the spatiotemporal information, which is to project the temporal dimension $\tau$ to the channel dimension $d_{\text{model}}$ by two convolutional layers with an in-between activation function.

\begin{figure}
  \begin{center}
    \includegraphics[width=\columnwidth]{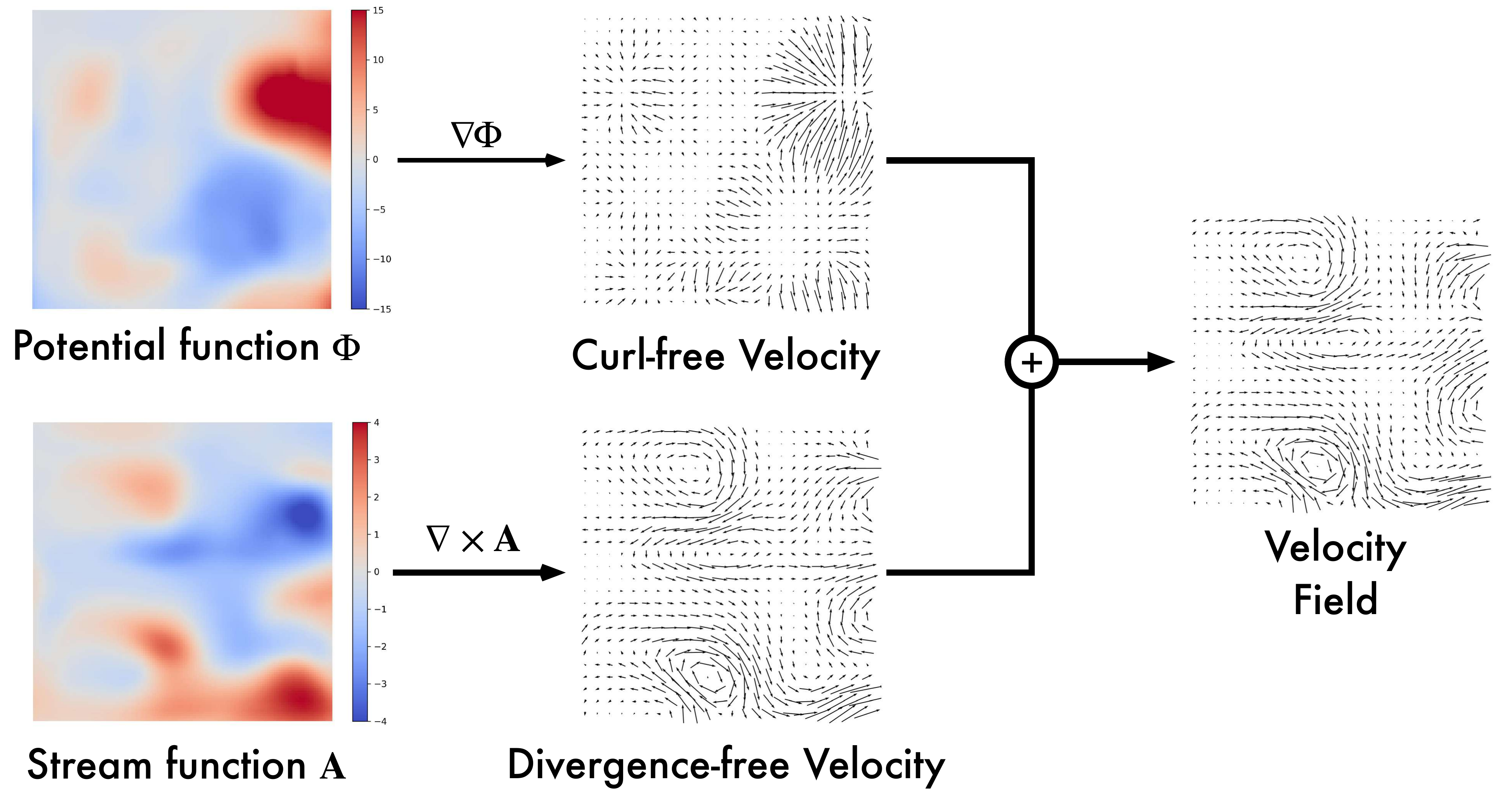}
  \end{center}
  \vspace{-15pt}
  \caption{{Transform potential and stream functions to velocity.}}\label{fig:vis_case}
  \vspace{-10pt}
\end{figure}

Next, following the convention in dynamics modeling \citep{sun2018pwc,teed2020raft}, we adopt \emph{spatiotemporal correlations} between fluid at the previous timestamp and the current timestamp to represent the dynamics information. Especially as physics quantities of fluid are highly affected by boundary conditions, we go beyond previous approaches and propose to further include boundary conditions $\mathbb{S}$ when calculating the spatiotemporal correlations:
\begin{equation}
\begin{aligned}
    \label{equ:approximate}
    \mathbf{c}(\mathbf{r}) &= \operatorname{Concat}\bigg(\big[ \widehat{\mathbf{x}}_{T}(\mathbf{r})\cdot \widehat{\mathbf{x}}_{T-1}(\mathbf{r}^\prime)\big]_{\mathbf{r}^\prime\in\mathbf{N}_{\mathbf{r}}},\\
    &\big[\textcolor{MyDarkRed}{\mathbbm{1}_{\mathbb{S}}(\mathbf{r}^\prime)}\left(\widehat{\mathbf{x}}_{T}(\mathbf{r})\cdot \widehat{\mathbf{x}}_{T-1}(\mathbf{r}^\prime)\right)\big]_{\mathbf{r}^\prime\in\mathbf{N}_{\mathbf{r}}}\bigg),\; \mathbf{r}\in\mathbb{V}
\end{aligned}
\end{equation}
where $\cdot$ denotes the inner-product operation and $\mathbf{N}_{\mathbf{r}}$ denotes the neighbors around position $\mathbf{r}$. $\mathbbm{1}_{\mathbb{S}}(\cdot)$ denotes the indicator function, whose value is 1 when $\mathbf{r}^\prime\in \mathbb{S}$ and 0 otherwise. $\mathbf{c}(\mathbf{r})\in\mathbb{R}^{2|\mathbf{N}_{\mathbf{r}}|}$ represents the correlation map between the current fluid at $\mathbf{r}$ and its $|\mathbf{N}_{\mathbf{r}}|$ neighbors in the previous fluid, with additional consideration on the boundary conditions $\mathbb{S}$. Thus, we obtain the extracted dynamics information $\mathbf{c}\in\mathbb{R}^{2|\mathbf{N}_{\mathbf{r}}|\times H\times W}$. Subsequently, we can decode the potential and stream functions from the dynamics information and calculate the Helmholtz dynamics as follows:
\begin{equation}
\begin{aligned}
    \label{equ:decoder}
    &\widehat{\Phi} = \operatorname{Decoder}_{\Phi}\left(\mathbf{c}\right),\ \widehat{\mathbf{A}} = \operatorname{Decoder}_{\mathbf{A}}\left(\mathbf{c}\right),\\
    &\widehat{\mathbf{F}}_{\text{Helm}}=\nabla\widehat{\Phi} + \nabla\times\widehat{\mathbf{A}},
\end{aligned}
\end{equation}
where $\widehat{\Phi},\widehat{\mathbf{A}}\in\mathbb{R}^{H\times W}$ and $\widehat{\mathbf{F}}_{\text{Helm}}\in\mathbb{R}^{2\times H\times W}$ represents the learned 2D fields of curl-free velocity, divergence-free velocity, and combined velocity respectively (Figure \ref{fig:vis_case}). $\operatorname{Decoder}_{\Phi}$ and $\operatorname{Decoder}_{\mathbf{A}}$ are learnable deep layers instantiated as two convolutional layers with an in-between activation function. We summarize the above process as
\begin{equation}
\widehat{\mathbf{F}}_{\text{Helm}}=\operatorname{HelmDynamics}(\hat{\mathbf{x}}_{(T-1)}, \hat{\mathbf{x}}_{T}).
\end{equation}
Note that we follow the standard practice in RAFT \citep{teed2020raft} and learn the fluid dynamics information from spatiotemporal correlations $\mathbf{c}$. However, rather than directly learning the velocity field from $\mathbf{c}(\mathbf{r})$ as in RAFT, we compose it from the learned potential and stream functions. Experimentally, this allows us to learn a more reasonable velocity field and enhances the accuracy of fluid prediction.

\begin{figure*}
\begin{center}
\centerline{\includegraphics[width=\textwidth]{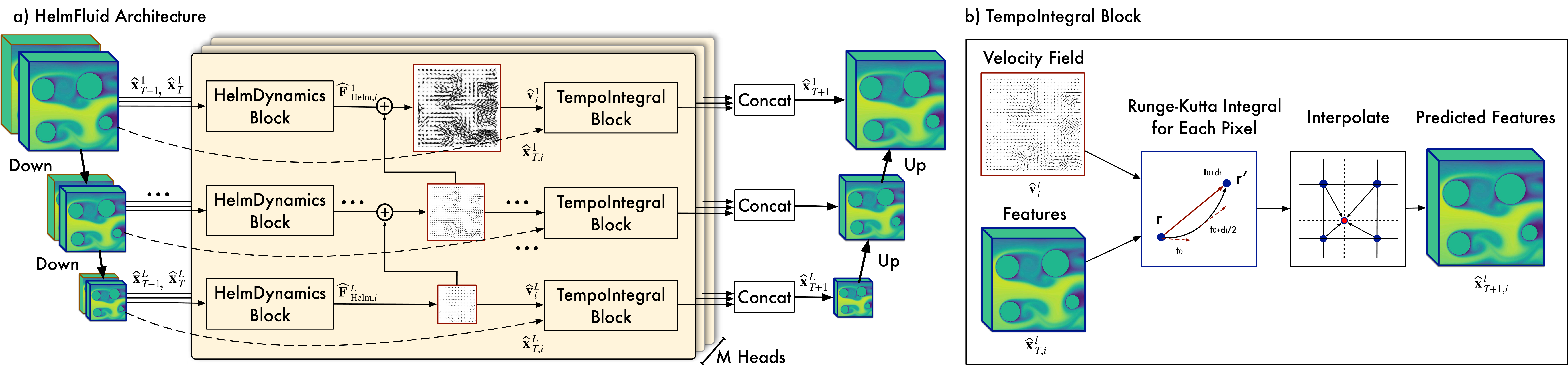}}
 \vspace{-5pt}
 \caption{HelmFluid architecture (left part), which employs Runge-Kutta with BFECC \citep{kim2005flowfixer} as a TempoIntegral Block to integrate the learned Helmholtz dynamics along the temporal dimension (right part) at multiple scales with multiheads to generate future fluid field. Especially, a residual connection across different scales is utilized to ensure the consistency of learned multiscale dynamics.}
    \label{fig:integration}
    \vspace{-20pt}
\end{center}
\end{figure*}

\subsection{Multiscale Multihead Integral Architecture}

After tackling intricate dynamics with the HelmDynamics blocks, we further present the Multiscale Multihead Integral Architecture to fuse learned dynamics along the temporal dimension for predicting future fluid, consisting of a multihead integral block and a multiscale modeling framework.

\vspace{-5pt}
\paragraph{Multihead dynamics} To predict the complex dynamics in fluid, we propose a multihead design for temporal integral, which is widely used in the attention mechanism to augment nonlinear capacity of deep models \citep{NIPS2017_3f5ee243}. The multihead design enables the model to capture different dynamic patterns via multiple Helmholtz dynamics $\widehat{\mathbf{F}}_{\text{Helm}, i}$ learned within different heads. As shown in Figure \ref{fig:integration}, given the deep representations $\widehat{\mathbf{x}}_{(T-1)},\widehat{\mathbf{x}}_{T}\in\mathbb{R}^{d_{\text{model}}\times H\times W}$ of two successive frames of fluid, we first split them into multiple heads along the channel dimension, with each head $i$ as $\widehat{\mathbf{x}}_{(T-1),i},\widehat{\mathbf{x}}_{T,i}\in\mathbb{R}^{\frac{d_{\text{model}}}{M}\times H\times W}, i\in\{1,\cdots,M\}$, where $M$ is the number of heads. Then we compute the Helmholtz dynamics from the deep representations for each head $i$:
\begin{equation}
    \label{equ:helmdynamic}
    \widehat{\mathbf{F}}_{\text{Helm}, i} = \operatorname{HelmDynamics}(\widehat{\mathbf{x}}_{(T-1),i}, \widehat{\mathbf{x}}_{T,i}),
\end{equation}
where $\widehat{\mathbf{F}}_{\text{Helm}, i}\in\mathbb{R}^{2\times H\times W}, i=1,\cdots,M$. 

\vspace{-5pt}
\paragraph{Multiscale modeling} It is known in physics that the fluid exhibits different properties at different scales. These multiscale dynamics entangle with each other, making the fluid extremely intractable. Thus, we adopt a multiscale modeling framework to enhance dynamics modeling.

Given input embeddings $\widehat{\mathbf{x}}_{(T-1)},\widehat{\mathbf{x}}_{T}\in\mathbb{R}^{d_{\text{model}}\times H\times W}$, we adopt a multiscale encoder to obtain deep representations in $L$ scales: $\widehat{\mathbf{x}}^l_{(T-1)},\widehat{\mathbf{x}}^l_{T}\in\mathbb{R}^{d^l_{\text{model}}\times\lfloor\frac{H}{2^{(l-1)}}\rfloor\times \lfloor\frac{W}{2^{(l-1)}}\rfloor}, l\in\{1,\cdots, L\}$. As the dynamics at larger scales are less affected by noise and more capable of giving a reliable background velocity field for the smaller scales, we ensemble the learned dynamics from coarse to fine to ease the multiscale dynamics modeling process. As shown in Figure \ref{fig:integration}, we obtain the velocity field $\widehat{\mathbf{v}}_i^l$ at the $l$-th scale by
\begin{equation}\label{equ:process}
    \widehat{\mathbf{v}}_i^l = 
    \begin{cases}
    \ \ \widehat{\mathbf{F}}_{\text{Helm}, i}^{l}, \ \ l=L\\
    \ \ \widehat{\mathbf{F}}_{\text{Helm},i}^{l} + \operatorname{Upsample}(\widehat{\mathbf{v}}_i^{l+1}), \ \ 1\leq l<L 
    \end{cases}
\end{equation}
where $\widehat{\mathbf{v}}_{i}^l\in\mathbb{R}^{2\times H\times W}$, and $\operatorname{Upsample}(\cdot)$ is the bilinear interpolation to keep resolution compatible. We incorporate the idea of residual learning for multiscale dynamics to align the velocity values. This ensures the consistency of velocity field at the fine scale with that at the coarse scale.

\vspace{-5pt}
\paragraph{TempoIntegral block} To predict the future fluid field, we integrate the feature space along the temporal dimension. Concretely, for scale $l \in {1,2,..., L}$ and head $i \in {1,2,..., M}$, we integrate the deep representation $\widehat{\mathbf{x}}_{T, i}^l$ by its corresponding velocity field $\widehat{\mathbf{v}}_{i}^l$. As shown in Figure \ref{fig:integration}, for a position $\mathbf{r}$, we take the second-order Runge-Kutta method \citep{devries1994first} as the numerical integral method to estimate its position in the future ${\rm d}t$ time: $\mathbf{r}_{i}^{l}{}^\prime = \mathbf{r} + \widehat{\mathbf{v}}_{i}^l(\mathbf{r}+\widehat{\mathbf{v}}_{i}^l(\mathbf{r})\frac{{\rm d}t}{2}){\rm d}t$. This equation can directly deduce the next step representation by moving the pixel at $\mathbf{r}$ to $\mathbf{r}^\prime$. Following the convention in temporal integral, we adopt the back-and-forth error compensation and correction (BFECC, \citep{kim2005flowfixer}) for better position mapping, which enhances the Runge-Kutta with an ensemble of bidirectional integral. Since the mapped coordinates $\mathbf{r}_{i}^{l}{}^\prime$ may not be integer positions on regular grids, we further use bilinear interpolation to yield representations on regular grids. We summarize the whole temporal integral process for each head $i$ at each scale $l$ as
\begin{equation}
\begin{aligned}
\label{equ:rk2}    \widehat{\mathbf{x}}_{(T+1),i}^l&=\operatorname{Interpolate}\left(\operatorname{BFECC}\big(\widehat{\mathbf{x}}_{T, i}^l,\widehat{\mathbf{v}}_{i}^l\big)\right)\\
\widehat{\mathbf{x}}_{(T+1)}^l&= \operatorname{Concat}\Big(\big[\widehat{\mathbf{x}}_{(T+1),i}^l\big]_{i=1,\cdots,M}\Big).
\end{aligned}
\end{equation}
 Eventually, we progressively aggregate the predicted context features from large to small scales and obtain the final prediction of the fluid field with a projection layer. More details of the model architecture including implementation of the BFECC method are deferred to Appendix \ref{appendix:implementation}.

\section{Experiments}\label{sec:exp}

\begin{figure*}
\begin{minipage}[!b]{0.63\textwidth}
 \hspace{-20pt}
\begin{center}
 \vspace{-5pt}
    \includegraphics[width=\textwidth]{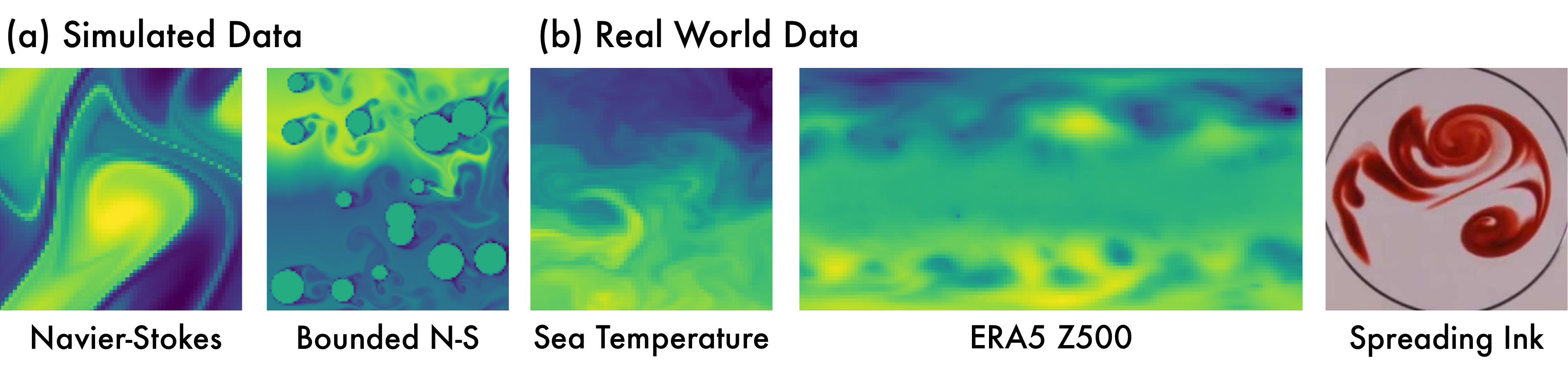}
  \end{center}
\end{minipage}
\hfill
 \begin{minipage}[!b]{0.35\textwidth}
\begin{center}
\begin{threeparttable}
\begin{sc}
\begin{small}
\setlength{\tabcolsep}{3pt}
\begin{tabular}{c|c|c}
\toprule
Benchmark & Boundary & Length\\
\midrule
Navier-Stokes  & Unknown & 10 \\ 
Bounded N-S & Known  & 10 \\ 
\midrule
ERA5 Z500 & Known  & 10 \\ 
Sea Temperature & Unknown  & 10 \\ 
Spreading Ink & Known  & 46-63 \\ 
\bottomrule
\end{tabular}
\end{small}
\end{sc}
\end{threeparttable}
\end{center}
\end{minipage}
\caption{Summary of five experiment benchmarks, including (a) simulated and (b) real-world data.}\label{fig:datasets} 
\vspace{-5pt}
\end{figure*}

\begin{table*}[t]
\vspace{-5pt}
    \caption{Performance comparison on the Navier-Stokes dataset under different resolutions. Relative L2 is recorded. For clarity, the best result is in bold and the second best is underlined. The relative promotion is calculated between the best and second-best models, that is $1-\frac{\text{The best error}}{\text{The second best error}}$. The timewise error curve is recorded from the $64\times 64$ settings.}
    \vspace{-5pt}
    \setlength{\tabcolsep}{3pt}
    \begin{minipage}[!b]{0.6\textwidth}
    \begin{table}[H]
    \begin{center}
    \begin{threeparttable}
    \begin{small}
    \begin{sc}
    \begin{tabular}{l|ccc}
    \toprule
    Model & {$64\times64$} & {$128\times 128$} & {$256\times 256$} \\
    \midrule
    DARTS \citep{ruzanski2011casa} & {0.8046} & {0.7002} & {0.7904} \\
    U-Net \citep{ronneberger2015u} & {0.1982} & {0.1589} & {0.2953}\\
    FNO \citep{li2021fourier} & {0.1556} &  {0.1028} & {0.1645} \\ 
    MWT \citep{Gupta2021MultiwaveletbasedOL} & {0.1586} & \underline{0.0841} & {\underline{0.1390}} \\
    U-NO \citep{rahman2023uno} & {\underline{0.1435}} & {0.0913} & {0.1392}\\ 
    LSM \citep{wu2023solving} & {0.1535} & {0.0961} & {0.1973} \\
    \midrule
    HelmFluid (Ours) & {\textbf{0.1261}} & \textbf{0.0807} & {\textbf{0.1310}}\\
    Promotion & 12.1\% & 4.0\% & 5.8\% \\ 
    \bottomrule
    \end{tabular}
    \end{sc}
    \end{small}
    \end{threeparttable}
    \end{center}
    \end{table}
    \end{minipage}
    \hfill
    \begin{minipage}[!b]{0.39\textwidth}
      \begin{center}
    \vspace{10pt}
    \hspace{-15pt}
    \includegraphics[width=0.99\textwidth]{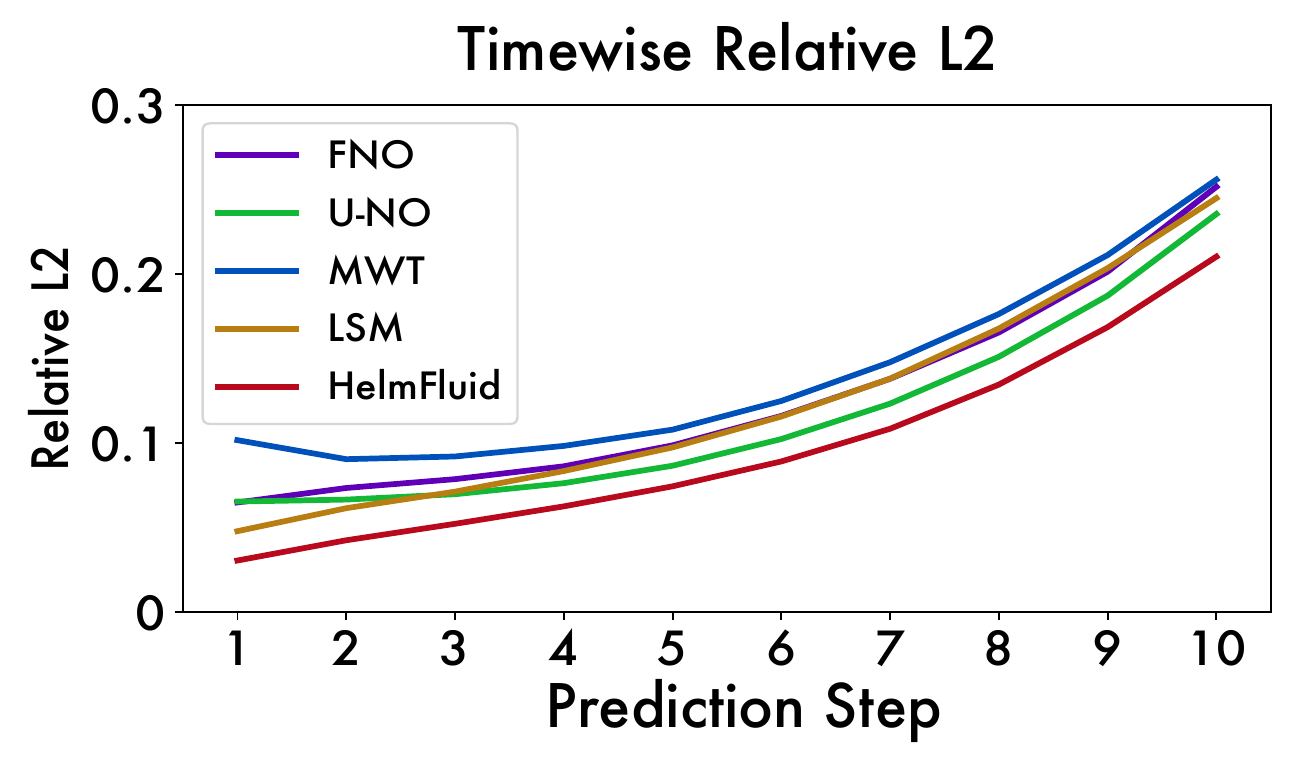}
  \end{center}
    \end{minipage}
    \label{nsresults}
    \vspace{-10pt}
\end{table*}

We extensively evaluate HelmFluid on five benchmarks, including both simulated and real-world observed scenarios, covering known and unknown boundary settings (see Figure~\ref{fig:datasets}). Extensions to 3D fluid are included in Appendix \ref{sec: 3DHelmFluid}.
Descriptions of datasets, baselines, and implementation details are listed in Appendix \ref{sec:implementation}.

\vspace{-5pt}
\paragraph{Baselines} We compare HelmFluid with nine competitive baselines, including one numerical method DARTS~\citep{ruzanski2011casa}, four neural fluid simulators: LSM~\citep{wu2023solving}, U-NO~\citep{rahman2023uno}, WMT~\citep{Gupta2021MultiwaveletbasedOL}, FNO~\citep{li2021fourier}, two fluid-dynamics-modeling solutions: Vortex~\citep{deng2023learning}, PWC-Net with fluid Refinement~\citep{zhang2022learning}, one vision backbone widely-used in fluid modeling: U-Net~\citep{ronneberger2015u}, and one model specialized for weather forecasting: FourcastNet~\citep{pathak2022fourcastnet}. Here, LSM and U-NO are previous state-of-the-art models in fluid prediction. Note that due to the inconsistent settings in fluid prediction, some of the baselines are not suitable for all benchmarks. Thus, in the main text, we only provide comparisons to baselines on their official benchmarks. But to ensure transparency, we also provide the complete results for other baselines in Table~\ref{tab:align_baseline}.

\subsection{Simulated Data}
\paragraph{Navier-Stokes with unknown boundary} This dataset is simulated from a viscous, incompressible fluid field on a two-dimensional unit torus, which obeys Navier-Stokes equations \citep{li2021fourier}. The task is to predict the future 10 steps based on the past 10 observations.

As presented in Table \ref{nsresults}, HelmFluid significantly surpasses other models, demonstrating its advancement in fluid prediction. In comparison with the second-best model, HelmFluid achieves 12.1\% relative error reduction (0.1261 vs.~0.1435) in the $64\times 64$ resolution setting and achieves consistent state-of-the-art in all time steps. Besides, HelmFluid performs best for the inputs under various resolutions, verifying its capability to handle the dynamics at different scales.

\begin{figure*}[t]
\begin{center}
\centerline{\includegraphics[width=1.0\textwidth]{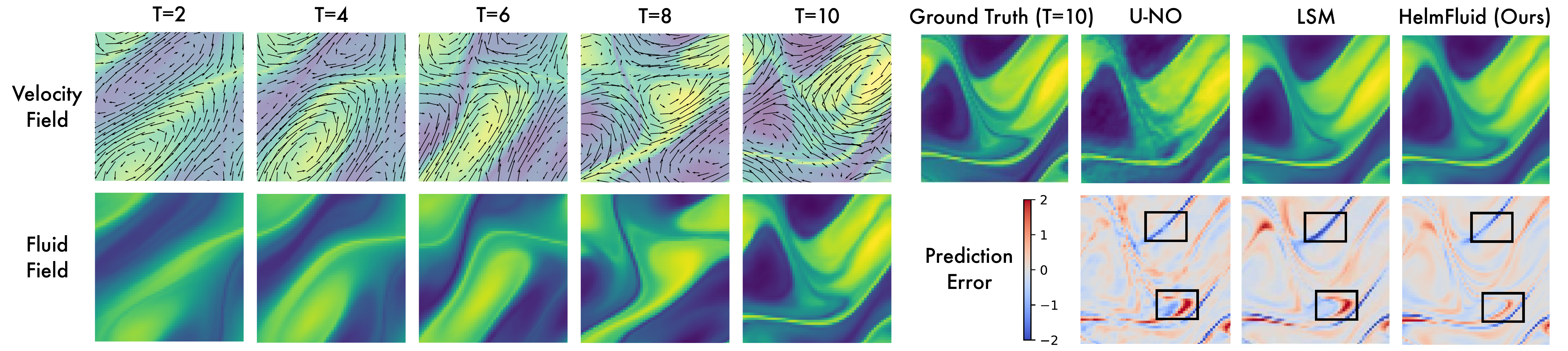}}
 \vspace{-5pt}
 \caption{\small{Showcase comparison and learned Helmholtz dynamics on the Navier-Stokes dataset under the $64\times 64$ input resolution.}}
    \label{fig:nsvelocityresults}
    \vspace{-15pt}
\end{center}
\end{figure*}

To intuitively present the model capability, we also provide several showcases in Figure~\ref{fig:nsvelocityresults}. In comparing to U-NO and LSM, HelmFluid precisely predicts the fluid motion, especially the twist parts, which involve complex interactions among several groups of fluid particles. Besides, HelmFluid also generates the learned velocity field for each step, which reflects the rotation and diffusion of fluid, empowering prediction with interpretable evidence. These results demonstrate the advantages of HelmFluid in capturing complex dynamics and endowing model interpretability.

\begin{figure*}[t]
\begin{center}
\centerline{\includegraphics[width=\textwidth]{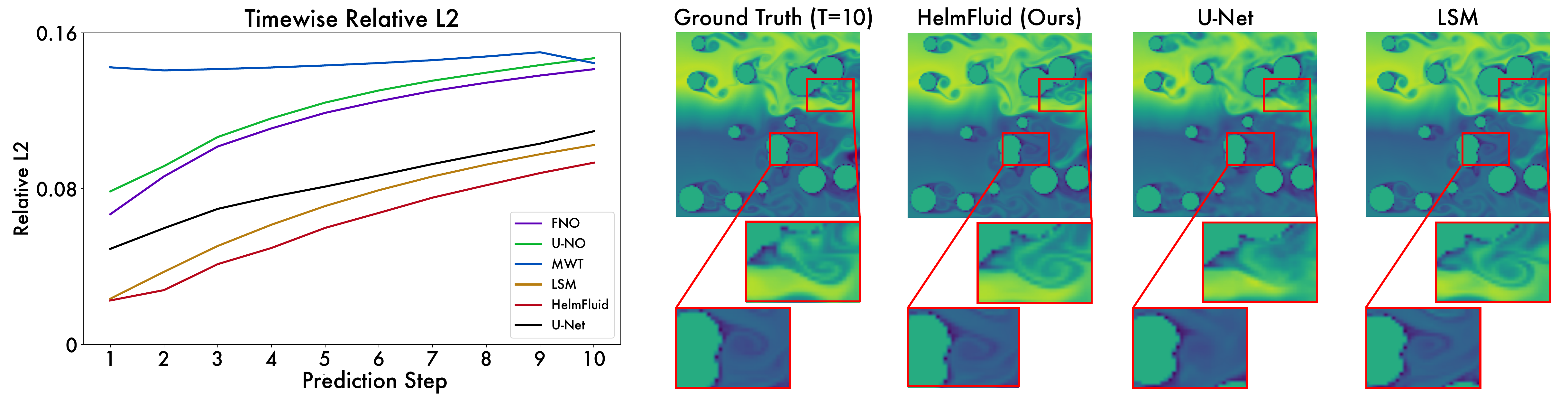}}
 \vspace{-10pt}
 \caption{{Timewise error and showcases on Bounded N-S dataset. For clarity, we highlight and zoom in the key parts of fluid in \textcolor{red}{red} boxes.}}
    \label{fig:boundedresults}
    \vspace{-20pt}
\end{center}
\end{figure*}

\begin{table}[t]
\vspace{-5pt}
\caption{Model performance comparison on Bounded N-S dataset.}
\label{tab:boundaryresults}
\begin{center}
\begin{threeparttable}
\begin{small}
\setlength{\tabcolsep}{8pt}
\begin{sc}
\begin{tabular}{l|cc}
    \toprule
    Model & Relative L2 \\
    \midrule
    DARTS \citep{ruzanski2011casa} & {0.1820} \\
    U-Net \citep{ronneberger2015u} & {
0.0846}\\
    FNO \citep{li2021fourier}& {0.1176} \\ 
    MWT \citep{Gupta2021MultiwaveletbasedOL} & {0.1407} \\
    U-NO \citep{rahman2023uno} & {0.1200}\\ 
    LSM \citep{wu2023solving} & {\underline{0.0737}} \\
    \midrule
    HelmFluid (Ours)& {\textbf{0.0652}}\\
    Promotion & 11.5\%\\
    \bottomrule
    \end{tabular}
\end{sc}
\end{small}
\end{threeparttable}
\end{center}
\vspace{-15pt}
\end{table}

\vspace{-5pt}
\paragraph{Bounded N-S with known boundary} This dataset simulates a wide pipe scenario, where the incompressible fluid moves from left to right, passing by several solid columns with position and size fixed in the dataset. The goal is to predict the future 10 steps based on past 10 observations.

HelmFluid also performs best in this challenging task and presents a consistent advantage in all prediction steps. While U-Net seems to be close to HelmFluid in averaged relative L2, it fails to capture the Karmen vortex phenomenon and results in blurry predictions (Figure \ref{fig:boundedresults}), which will seriously impede its interpretability. In contrast, HelmFluid precisely predicts the Karmen vortex around boundaries with eidetic texture. This result benefits from the learning paradigm designed based on Helmholtz dynamics.

Besides, we provide the comparison of learned dynamics among HelmFluid, DARTS \citeyearpar{ruzanski2011casa} and PWC-Net \citeyearpar{zhang2022learning} in Figure \ref{fig:introduction}. HelmFluid shows impressive capturing of the dynamics accurately, even for vortices around solid columns. This capability stems from HelmFluid's design in learning potential and stream functions instead of directly learning velocities, thereby mitigating overwhelming the model by intricate dynamics. It is notable that the numerical method DARTS degenerates seriously in both quantitative results (Table \ref{tab:boundaryresults}) and learned dynamics (Figure \ref{fig:introduction}), which highlights challenges in this task and the advantage of HelmFluid.

\begin{table}[t]
    \vspace{-10pt}
    \caption{Model comparison on the ERA5 Z500 dataset.}
    \vspace{3pt}
    \label{tab:era5results}
    \begin{center}
    \begin{threeparttable}
    \begin{small}
    \setlength{\tabcolsep}{10pt}
    \begin{sc}
    \begin{tabular}{l|cc}
    \toprule
    Models& RMSE\\
    \midrule
    U-Net \scalebox{0.9}{{\citep{ronneberger2015u}}} &  {632.94}\\
    FNO \scalebox{0.8}{{\citep{li2021fourier}}} &  {596.80} \\ 
    MWT \scalebox{0.9}{{\citep{Gupta2021MultiwaveletbasedOL}}} &  {596.45} \\
    U-NO \scalebox{0.9}{{\citep{rahman2023uno}}} & {596.84 } \\ 
    LSM \scalebox{0.8}{{\citep{wu2023solving}}} &  \underline{561.27} \\
    FourcastNet \scalebox{0.9}{{\citep{pathak2022fourcastnet}}} &  {594.49} \\
    \midrule
    HelmFluid (Ours) & \textbf{521.44} \\
    Promotion & 7.1\% \\
    \bottomrule
    \end{tabular}
    \end{sc}
    \end{small}
    \end{threeparttable}
    \end{center}
    \vspace{-20pt}
    \end{table}

\vspace{-5pt}
\subsection{Real-world Data}
\paragraph{ERA5 Z500 with known boundary} \label{par: ERA5 Z500 dataset} This dataset is processed from the fifth generation of ECMWF reanalysis (ERA5) data \citep{hersbach2020era5}. Following the practice of weather forecasting, we choose the geopotential height at 500 hPa (Z500) with a resolution of $2.5^{\circ}$ and a time interval of 3 hours as our data. The goal is to predict the geopotential height for 10 timesteps given 2 observations.

The Root Mean Square Error (RMSE) for Z500 is detailed in Table \ref{tab:era5results}, revealing that HelmFluid consistently outperforms all other comparative baselines. Notably, it surpasses FourcastNet \citeyearpar{pathak2022fourcastnet},
the pioneering model to leverage ERA5 reanalysis data for medium-range meteorological forecasts. This superior performance indicates that HelmFluid is adept at predicting real-world atmospheric fluid dynamics and has the potential to advance weather forecasting capabilities. More showcases can be found in Figure \ref{fig:z500showcase}.

 \begin{table}[t]
    \vspace{-10pt}
    \caption{Model comparison on the Sea Temperature dataset.}
    \label{tab:searesults}
    \begin{center}
    \begin{threeparttable}
    \begin{small}
    \setlength{\tabcolsep}{3pt}
    \begin{sc}
    \begin{tabular}{l|cc}
    \toprule
    Models& Relative L2 & MSE   \\
    \midrule
    DARTS \scalebox{0.8}{{\citep{ruzanski2011casa}}} & {0.3308} & {0.1094} \\
    U-Net \scalebox{0.8}{{\citep{ronneberger2015u}}} & \underline{0.1735} & \underline{0.0379}\\
    FNO \scalebox{0.8}{{\citep{li2021fourier}}} & {0.1935} & {0.0456} \\ 
    MWT \scalebox{0.8}{{\citep{Gupta2021MultiwaveletbasedOL}}} & {0.2075} & {0.0506} \\
    U-NO \scalebox{0.8}{{\citep{rahman2023uno}}} & {0.1969} & {0.0472} \\ 
    LSM \scalebox{0.8}{{\citep{wu2023solving}}} & {0.1759} & {0.0389} \\
    \midrule
    HelmFluid (Ours) & \textbf{0.1704} & \textbf{0.0368} \\
    Promotion & 1.8\% & 2.9\% \\
    \bottomrule
    \end{tabular}
    \end{sc}
    \end{small}
    \end{threeparttable}
    \end{center}
    \vspace{-15pt}
    \end{table}

\begin{figure*}[t]
\begin{center}
\vspace{-5pt}
\centerline{\includegraphics[width=\textwidth]{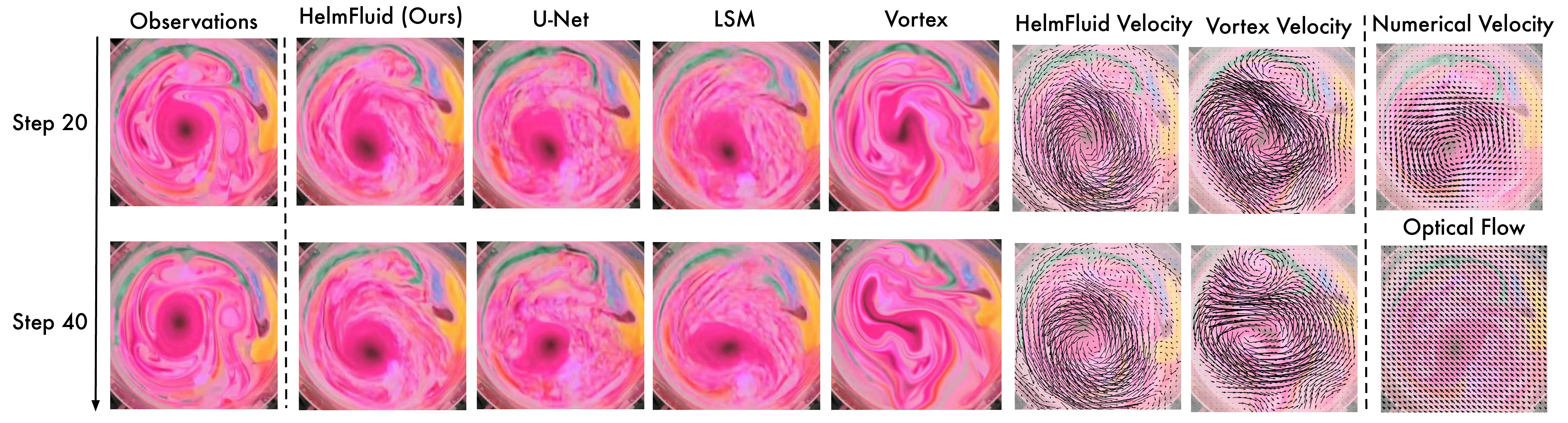}}
 \vspace{-10pt}
 \caption{Showcases of prediction results (future 20 and 40 steps) and learned velocity fields (future 40 steps) on the Spreading Ink dataset. We also show velocity fields of the first timestep estimated by numerical method (DARTS) and deep optical flow (PWC-Net).} 
    \label{fig:vortex}
    \vspace{-20pt}
\end{center}
\end{figure*}

\vspace{-5pt}
\paragraph{Sea Temperature with unknown boundary} \label{par: sea temperature dataset} This dataset consists of the reanalysis ocean temperature data \citeyearpar{oceanreanalysis} provided by ECMWF. We adapted the data in four $64\times 64$ regions located in different oceans. The goal is to predict the temperature for the next 10 days based on 10 past days.

The results in Table \ref{tab:searesults} demonstrate that HelmFluid can handle real-world data well and outperform all baselines. It is worth noting that the test set is collected from different regions with respect to the training and validation sets, which involves the distribution shift problem. Thus, these results also verify the generality and transferability of HelmFluid.

\vspace{-5pt}
\paragraph{Spreading Ink with known boundary}
This benchmark consists of three videos collected by \citeauthor{deng2023learning}, involving more than 100 successive frames respectively. Following the experiment setting in Vortex \citeyearpar{deng2023learning}, the goal is to predict the last 1/3 frames of the video using the first 2/3.

\begin{table}[t]
\vspace{-10pt}
\caption{Model comparison on Spreading Ink. Averaged Perceptual loss, Relative L2 and MSE of three sub-videos are reported.}
\label{tab:realresults}
\vspace{-10pt}
\begin{center}
\begin{threeparttable}
\begin{small}
\begin{sc}
\begin{tabular}{l|ccc}
\toprule
Models & Metrics \\
\midrule
U-Net \scalebox{0.8}{\citep{ronneberger2015u}} & \underline{3.596} / 0.2620	/ 0.0176 \\
FNO \scalebox{0.8}{\citep{li2021fourier}}& 4.095 / 0.2776 / 0.0198 \\ 
U-NO \scalebox{0.8}{\citep{rahman2023uno}} & 5.604 / 0.2971 / 0.0227\\
Vortex \scalebox{0.8}{\citep{deng2023learning}} & 3.949 / \underline{0.2483} / \underline{0.0161} \\
LSM \scalebox{0.8}{\citep{wu2023solving}}  & 3.760 / 0.2698 / 0.0187 \\ 
\midrule
HelmFluid (Ours) & \textbf{3.323} / \textbf{0.2183} / \textbf{0.0125}\\
Promotion & 7.6\% / 12.1\% / 22.3\%\\
\bottomrule
\end{tabular}
\end{sc}
\end{small}
\end{threeparttable}
\end{center}
\vspace{-15pt}
\end{table}

The quantitative results are listed in Table \ref{tab:realresults}. HelmFluid still performs well in this long-term forecasting task. In addition to the relative L2 and MSE, it also consistently achieves the lowest VGG perceptual loss, implying that the prediction results of HelmFluid can maintain the realistic texture and intuitive physics. As for showcases in Figure \ref{fig:vortex}, we find that HelmFluid can precisely capture the diffusion of ink. Even for the future 40 frames, HelmFluid still performs well in capturing the hollow position and surpasses numerical methods, optical flow and Vortex, in learning the velocity.

\subsection{Model analysis}

\begin{figure}[b]
\vspace{-10pt}
    \centering
    \includegraphics[width=\columnwidth]{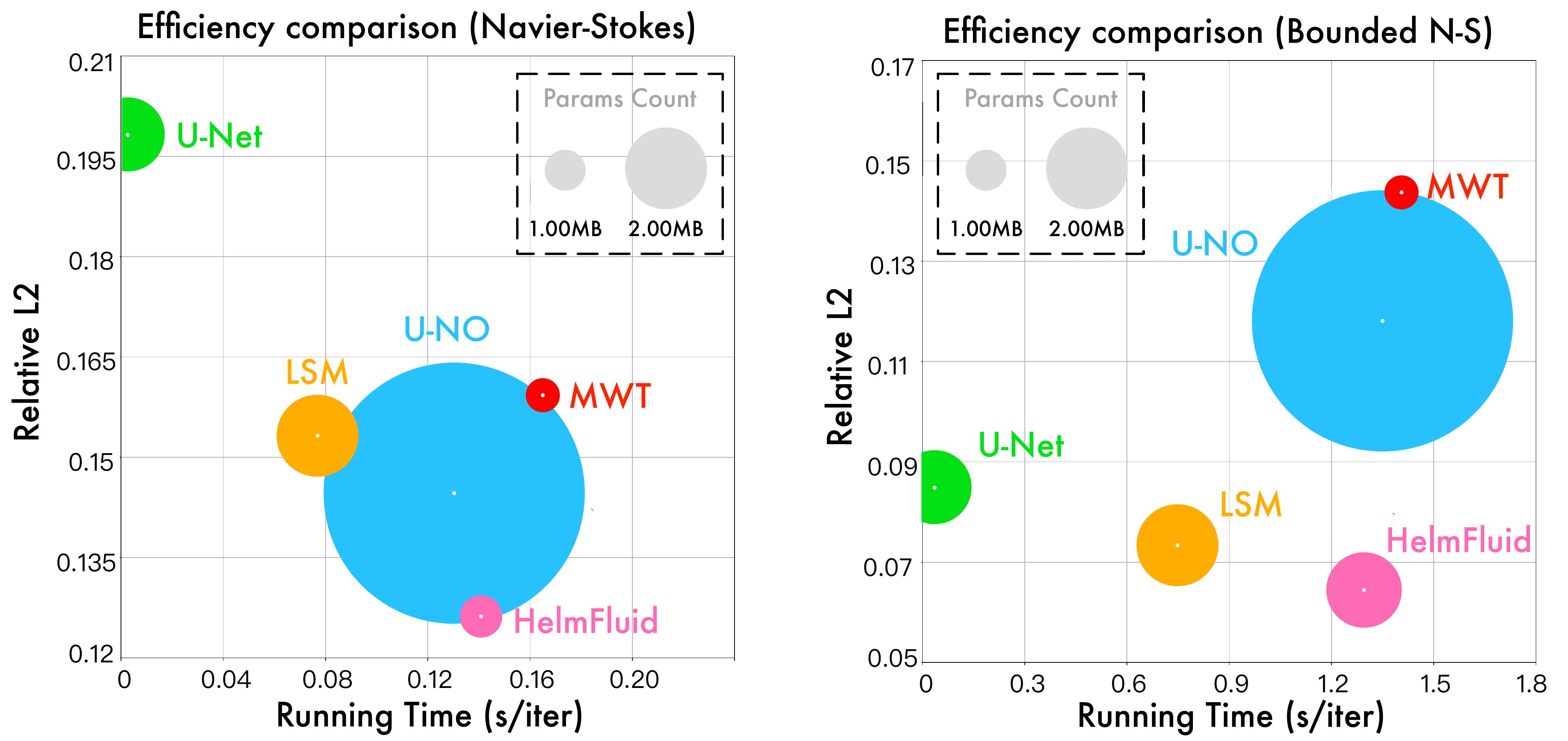}
    \vspace{-20pt}
    \caption{Efficiency comparison evaluated on the $64 \times 64$ Naiver-Stokes and $128 \times 128$ Bounded N-S averaged from $10^3$ iterations.}\label{fig:efficiency}
\end{figure}

\paragraph{Efficiency analysis} 
To evaluate model practicability, we also provide efficiency analysis in Figure \ref{fig:efficiency}. In comparison with the second-best model U-NO, HelmFluid presents a favorable trade-off between efficiency and performance. In particular, HelmFluid surpasses U-NO by 12.1\% in relative L2 with comparable running time. See Appendix \ref{sec:additional_results} for full results and comparisons under aligned model size.

\begin{table}[t]
\vspace{-10pt}
\caption{Ablation on the HelmDynamics block, which includes learning w/o HelmDynamics on the $64\times 64$ Navier-Stokes dataset, and learning w/o boundary conditions on Bounded N-S dataset.}
\label{tab:ablation_main}
\begin{center}
\begin{threeparttable}
\setlength{\tabcolsep}{2.2pt}
\begin{small}
\begin{sc}
\begin{tabular}{l|l|c}
\toprule
Data & Model & Relative L2\\
\midrule
Navier-& Directly Learning Velocity & 0.1412 \\
Stokes& Learning HelmDynamics & \textbf{0.1261} \\ 
& Promotion & 10.7\%\\
\midrule
Bounded& w/o Boundary Conditions & 0.0846 \\
N-S & w/ Boundary Conditions & \textbf{0.0652} \\ 
& Promotion & 22.9\%\\
\bottomrule
\end{tabular}
\end{sc}
\end{small}
\end{threeparttable}
\end{center}
\vspace{-15pt}
\end{table}

\begin{figure*}[t]
\begin{center}
\centering
    \includegraphics[width=\textwidth]{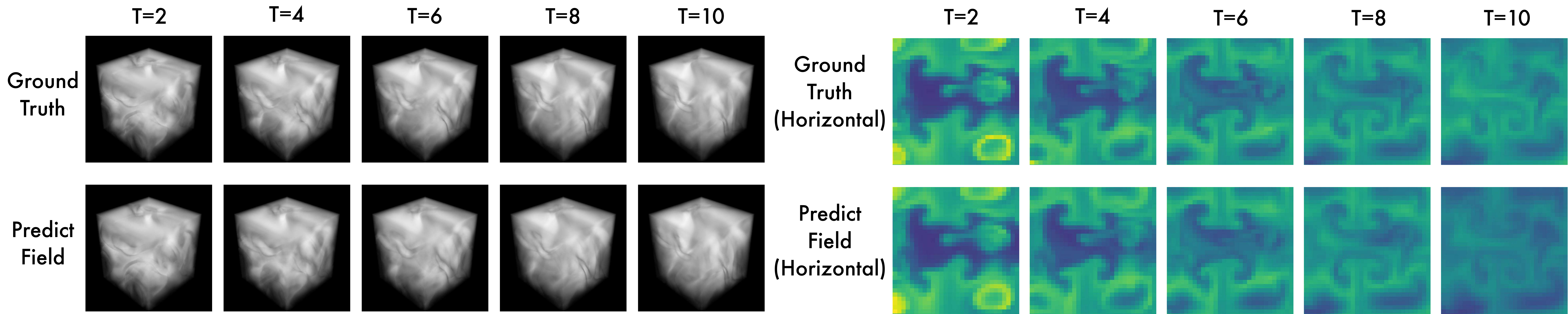}
\vspace{-15pt}
    \caption{Showcases of the 3D Smoke Dataset, visualization from 3D perspective and 2D horizontal slice are both provided.}\label{fig:smoke_data}
\vspace{-10pt}
\end{center}
\end{figure*}

\vspace{-5pt}
\paragraph{Ablations} To highlight advantages of learning Helmholtz dynamics, we compare HelmFluid with its two variants: directly learning the velocity field and removing the boundary condition design. As shown in Table \ref{tab:ablation_main}, directly estimating the velocity field will cause 10.7\% performance drop. A plausible reason is that deep models can be overwhelmed by complex fluid interactions, and thus learning Helmholtz dynamics is beneficial. Also, our design in incorporating boundary condition is essential. Empowered by the HelmDynamics block, our model can conveniently utilize the boundary information, unleashing its potential in handling fluid with complex boundaries. Complete quantitative and visual comparisons are included in Appendices \ref{sec:ablation} and \ref{section:showcases}.

\begin{figure}[tbp]
\centering
    \includegraphics[width=\columnwidth]{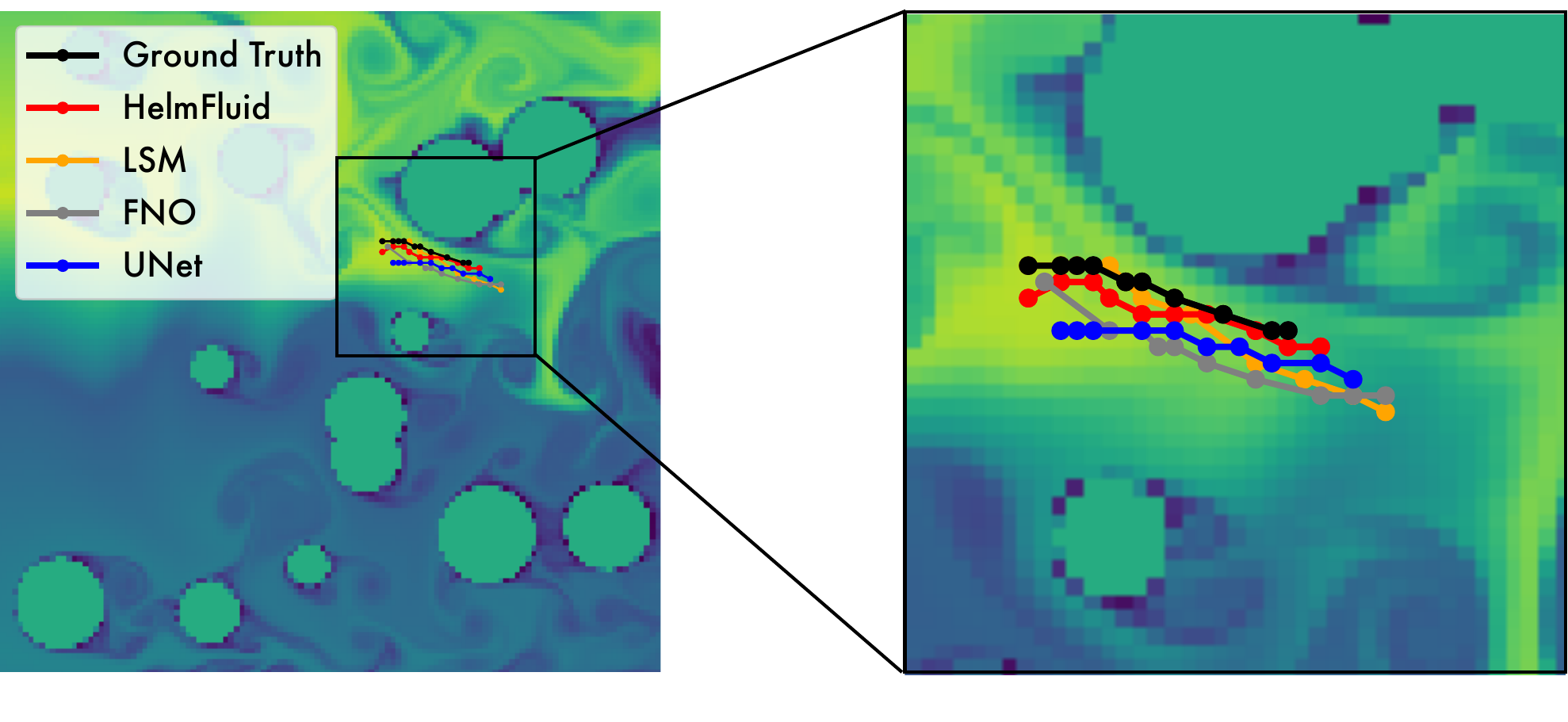}
    \vspace{-20pt}
    \caption{\small{Tracking of the local maxima of Bounded N-S.}}\label{fig:Tracking}
    \vspace{-10pt}
\end{figure}

\vspace{-5pt}
\paragraph{Dynamics tracking}
This paper is based on the Eulerian specification of fluid. As a supplement, we also provide a Lagrangian perspective to analyze model predictions. Technically, we locate the local maxima of fluid and keep tracking it in the subsequent frames. The closer the predicted trajectory of the point is to its real counterpart, the better tracking of the point is indicated. As shown in Figure \ref{fig:Tracking}, the trajectory predicted by HelmFluid is the closest to the ground truth, verifying the capability of HelmFluid in capturing the intricate dynamics of fluid.

\vspace{-5pt}
\paragraph{Generalization on boundary conditions}
HelmFluid incorporates boundary conditions as additional data for correlation calculations, enhancing its predictive capabilities. To evaluate the model's performance under various boundary conditions, we undertook a modification of the Bounded Navier-Stokes dataset by altering the geometric form of the solid columns from \emph{circular} to \emph{square}, thereby introducing a distinct set of boundary conditions for testing. We tested on three different training-testing scenarios, including zero-shot testing on the alternative dataset, fine-tuning the model parameters on the alternative dataset followed by testing, and blending two distinct datasets for separate testing. All the models are trained with the same number of epochs.

As shown in Table \ref{tab:generalization}, direct zero-shot testing on a new dataset leads to a significant degradation in model performance. However, merely fine-tuning for 3 epochs achieved comparable or even superior test results on the original dataset. In the scenario of mixed training, test metrics on both datasets improve by 2.9\% and 16.7\%, respectively. This proves the generalizability of HelmFluid in handling various boundary conditions and demonstrates an improvement as the dataset grows in size and diversity.

\begin{table}[t]
\vspace{-7pt}
\caption{Experimental results for generalized boundary conditions, \textit{circle} denotes the original Bounded N-S dataset, while \textit{square} denotes the modified dataset. The underlined metric indicates training performance with the test set being identical.}
\label{tab:generalization}
\begin{center}
\begin{threeparttable}
\setlength{\tabcolsep}{5pt}
\begin{small}
\begin{sc}
\begin{tabular}{l|c|c}
\toprule
Dataset & Circle & Square \\
\midrule
Trained on Circle &	\underline{0.0731} & 0.1501 \\ 
Fintuned 3 epochs on Circle & - & 0.0791 \\
Fintuned 100 epochs on Circle & - & 0.0781 \\
\midrule
Trained on Square & 0.1503 & \underline{0.0855} \\
Fintuned 3 epochs on Square & 0.0710 & - \\
Fintuned 100 epochs on Square & 0.0691 & - \\
\midrule
Trained on Circle and Square & \textbf{0.0618} & \textbf{0.0712} \\
\bottomrule
\end{tabular}
\end{sc}
\end{small}
\end{threeparttable}
\end{center}
\vspace{-26pt}
\end{table}

\subsection{Extend HelmFluid to 3D Fluid}
The form of Helmholtz decomposition implies its applicability across arbitrarily high dimensions. To address fluid prediction in three-dimensional scenarios, we extend the HelmFluid model to 3D and experiment on the simulated 3D smoke buoyancy dataset with known boundary. As shown in Figure \ref{fig:smoke_data}, HelmFluid generates smoke that closely matches the shape and position of the ground truth, effectively reflecting the variations in location and intensity of the smoke. See Appendix \ref{sec: 3DHelmFluid} for details of the dataset and implementation.

\section{Conclusions and Future Work}
In this paper, we present the HelmFluid model towards accurate and interpretable fluid prediction. Instead of directly learning velocity fields, we propose to learn the Helmholtz dynamics, which casts the intricate dynamics of fluid into inherent physics quantities. With HelmDynamics blocks and Multiscale Multiscale Integral Architecture, HelmFluid can precisely estimate the potential and stream functions for Helmholtz dynamics, which empowers the prediction process with physical interpretability. HelmFluid achieves consistent state-of-the-art on both simulated and real-world datasets, even for scenarios with complex boundaries. In the future, we plan to further extend HelmFluid to large-scale datasets, such as world climate and ocean current modeling. 

\vspace{-5pt}
\section*{Acknowledgements}
This work was supported by the National Key Research and Development Plan (2021YFC3000905), the National Natural Science Foundation of China (U2342217 and 62022050), the BNRist Innovation Fund (BNR2024RC01010), and the National Engineering Research Center for Big Data Software.

\section*{Impact Statement}

This paper presents work whose goal is to advance deep learning research for fluid prediction, which has the potential application to enhance the interpretability of atmospheric forecasts, ocean forecasts, and turbulence predictions in machine learning. Our work only focuses on the scientific problem, so there is no potential ethical risk.

\bibliography{example_paper}
\bibliographystyle{icml2024}

\appendix
\onecolumn

\section{Extend HelmFluid to 3D Fluid}
\label{sec: 3DHelmFluid}
Here we present the potential extension of HelmFluid to 3D fluid prediction. According to the formalization of Helmholtz decomposition ${\mathbf{F}}(\mathbf{r}) = \nabla\Phi(\mathbf{r}) + \nabla \times {\mathbf{A}}(\mathbf{r}), \mathbf{r} \in \mathbb{V}$. For 2D cases in the main text, the velocity component on the $z$-axis is set to be zero, thereby $\mathbf{A}_x(\mathbf{r})=\mathbf{A}_y(\mathbf{r})=0$. By extending the HelmDynamics block to learn $\widehat{\Phi}\in\mathbb{R}^{1\times D \times H\times W}$ and $\widehat{\mathbf{A}}\in\mathbb{R}^{3\times D \times H\times W}$, where $D$ is the additional depth dimension of 3D fluid, we can adapt HelmFluid to 3D fluid prediction. Then, following the Helmholtz decomposition presented in Eq.~\ref{equ:helmholtz1}, we can easily obtain the inferred 3D vector velocity field, thereby enabling HelmFluid to achieve the velocity-aware 3D fluid prediction.

\begin{figure*}[h]
\centering
    \includegraphics[width=\textwidth]{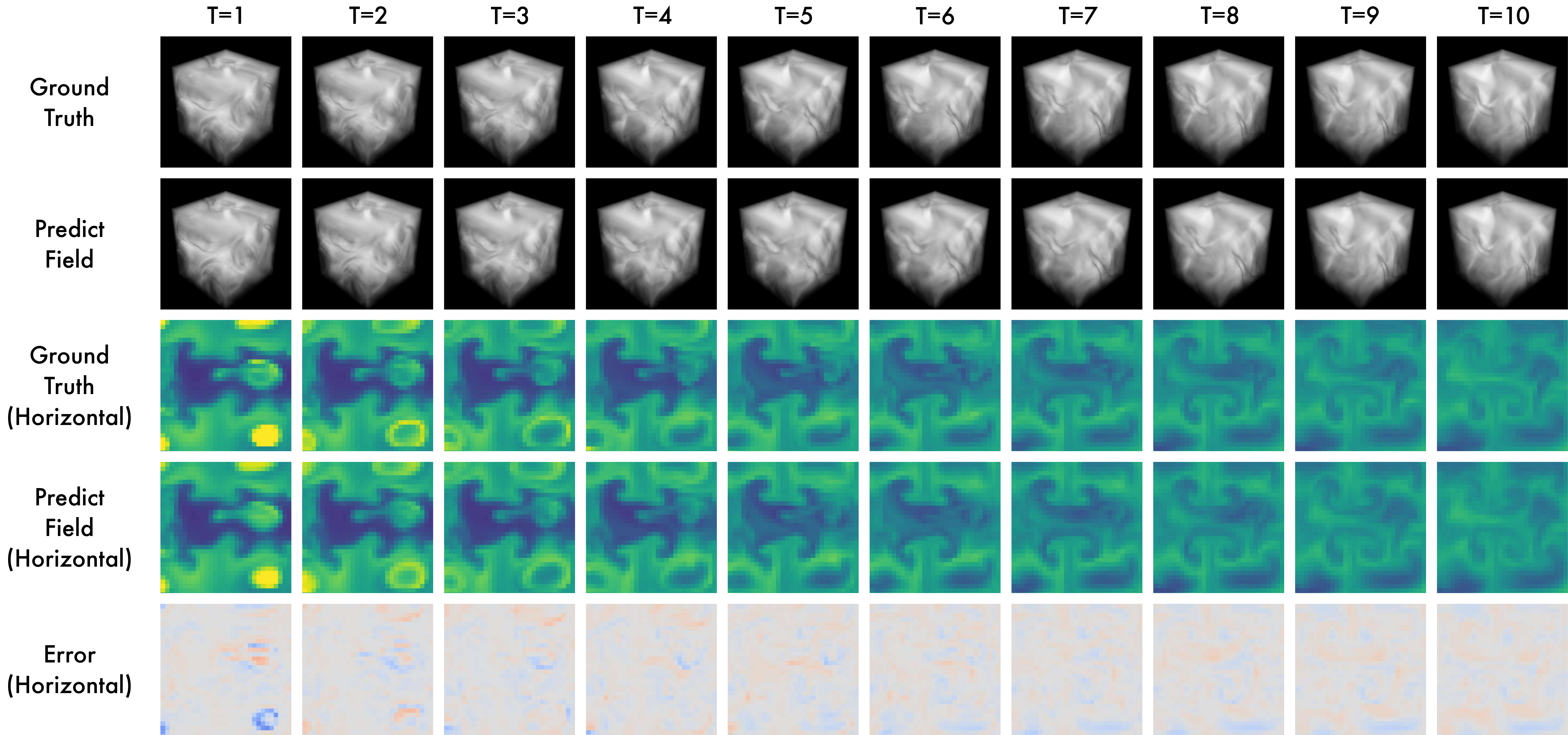}
    \caption{{Showcases of the 3D Dataset, visualization from 3D perspective and 2D horizontal slice are provided.}}\label{fig:smoke}
\end{figure*}

To validate the capabilities of HelmFluid on 3D fluid prediction scheme, we generated 3D smoke buoyancy dataset by modifying the 3D solver (under MIT license) from \href{https://github.com/BaratiLab/FactFormer/}{https://github.com/BaratiLab/FactFormer/}. 3D smoke buoyancy problem is governed by the incompressible Navier-Stokes equation coupled with advection equation \citep{li2023scalable}:
\begin{equation}
\begin{aligned}
\frac{\partial \mathbf{u}(\mathbf{x}, t)}{\partial t}+\mathbf{u}(\mathbf{x}, t) \cdot \nabla \mathbf{u}(\mathbf{x}, t) & =\nu \nabla^2 \mathbf{u}(\mathbf{x}, t)-\frac{1}{\rho} \nabla p(\mathbf{x}, t)+\mathbf{f}(\mathbf{x}, t), & & \mathbf{x} \in(0, L)^3, t \in(0, T], \\
\frac{\partial d(\mathbf{x}, t)}{\partial t}+\mathbf{u}(\mathbf{x}, t) \cdot \nabla d(\mathbf{x}, t) & =0, & & \mathbf{x} \in(0, L)^3, t \in(0, T], \\
\nabla \cdot \mathbf{u}(\mathbf{x}, t) & =0, & & \mathbf{x} \in(0, L)^3, t \in[0, T], \\
\mathbf{u}(\mathbf{x}, 0)=0, \quad \mathbf{d}(\mathbf{x}, 0) & =d_0(\mathbf{x}), \quad \mathbf{f}(\mathbf{x}, t) = \left[0, 0, \eta d(\mathbf{x}, t)\right] & & \mathbf{x} \in(0, L)^3, t \in(0, T],
\end{aligned}
\end{equation}
where $L$ is set $32$, $\eta$ is the buoyancy factor. The goal is to predict the future 10 steps based on the past 10 frames. We generated 1000 trajectories for training and 200 for testing and provided 3D showcases and 2D slices in Figure \ref{fig:smoke}.

\section{Implementation Details}\label{sec:implementation}

\subsection{Dataset}
\label{subsection:dataset}
We summarize the experiment datasets in Table \ref{tab:dataset}. More details can be found in the following.

\begin{table}[h]
\vspace{-5pt}
    \caption{A summary of experiment datasets. Note that the Spreading Ink dataset is different from other benchmarks, which only contains three video sequences. We strictly follow the Vortex \citep{deng2023learning} to split the video for training, validation and test. For example, in the training phase of video 1, we use the first 70 frames for training and the subsequent 30 frames for validation. As for the test, we use the first 100 frames as input and predict the following 50 frames.}
    \label{tab:dataset}
    \vskip 0.1in
    \centering
 \vspace{-5pt}
    \begin{small}
        \begin{sc}
            \renewcommand{\multirowsetup}{\centering}
            \setlength{\tabcolsep}{3pt}
            \begin{tabular}{l|c|c|c|c}
                \toprule
                Dataset & (Input, predict length) & (Training, validation, test) & Observed state & Reynold numbers \\
                \midrule
             Navier-Stokes & (10,10) & (1000,200,200) & Vorticity & $\sim 10^4$\\
              Bounded N-S & (10,10) & (1000,200,200) & Grayscale & $\sim 300$\\
              ERA5 Z500 & (2,10) & (20425, 2087, 4174) & Geopotential & Unknown \\
              Sea Temperature & (10,10) & (170249, 17758, 65286) & Temperature & Unknown\\
              Spreading Ink video 1 & (100, 50) & One video sequence & RGB Image & Unknown\\
              Spreading Ink video 2 & (126, 63) & One video sequence & RGB Image & Unknown\\
              Spreading Ink video 3 & (93, 46) & One video sequence & RGB Image & Unknown\\
                \bottomrule
            \end{tabular}
        \end{sc}
    \end{small}
 \vspace{-10pt}
\end{table}

\paragraph{Navier-Stokes} Navier-Stokes equations describe the motion of a viscous incompressible field. In this paper, we follow \citep{li2021fourier} and generate fluid on a 2D torus with the following equation:
\begin{equation}
\begin{aligned}
\label{NavierStokes}
\frac{\partial \mathbf{u}}{\partial t}+(\mathbf{u} \cdot \nabla) \mathbf{u}-\nu \nabla^2 \mathbf{u}&=-\frac{1}{\rho} \nabla p+\mathbf{g}\\
\nabla \cdot \mathbf{u} &= 0\\
\nabla \times \mathbf{u}(x,0) &= \omega_0(x), x\in (0,1)^2, \\
\end{aligned}
\end{equation}
where $\mathbf{u}\in \mathbb{R}^2$ represents the velocity field, $p$ denotes the pressure, and $\rho$ is the fluid density, which we assumed to be constant in the incompressible fluid field. $\nu \in \mathbb{R}_+$ is the kinematic viscosity representing the intrinsic nature of the fluid, which is assumed to be constant. $\mathbf{g}\in \mathbb{R}^2$ represents the summation of all the external forces applied on the fluid field. Vorticity is calculated by the velocity field, $\omega = \nabla \times \mathbf{u}$. At time zero, the initial vorticity field $\omega_0$ is given. The goal is to predict the following vorticity fields from given observations.

We randomly sampled the initial vorticity $w_0$ on a two-dimensional unit torus from a Gaussian distribution and solved the equation with a numerical method to obtain the future velocity field. After generating the fluid field with $256\times256$ spatial resolution and $10^{-4}$ second temporal resolution, we downsampled it to a sequence of 1 second per frame and corresponding spatial resolution. Thus, each sequence consists of 20 frames with a total duration of 20 seconds. We fixed the viscosity $\nu=10^{-5}$ for all three sub-datasets of different resolutions. To verify the model capacity in different resolutions, we generate three subsets ranging from $64\times 64$ to $256\times 256$ with 1000 training sequences, 200 validation sequences and 200 test sequences. The goal is to predict the future 10 steps based on the past 10 observations.

\paragraph{Bounded N-S}
In real-world applications, we usually need to handle the complex boundary conditions in fluid prediction. Specifically, suppose a $512\times512$ sized two-dimensional space with top and bottom as boundaries, with free space outside the image. We let a randomly colored fluid flow from left to right. To test the model performance under scenarios with complex boundaries, we randomly sampled fifteen circles of different sizes as obstacles and uniformly placed them in the $512\times512$ space. Then we used Taichi \citep{hu2019taichi} as a simulator engine to generate fluid within top-down boundaries with a numerical fluid solver for the advection equation \citep{baukal2000computational}, and generated a sufficiently long sequence with one initial source condition. The generated fluid field contains the Karmen vortex phenomenon \citep{bayindir2021efficient} with many vortices of various sizes, making this problem extremely challenging.

Towards the flow field dataset, after the colored fluid field spreads over the space from left to right, we sample frames in the frequency of 60 steps and add the sampled frames into the dataset. To remove the noise from chromatic aberration, we transform all the samples into grayscale. Then we downsample the image to $128\times128$ and split the long sequence into disjoint subsequences with 20 timesteps.  After randomly dividing them into train, validation, and test sets, we finally obtained the training, validation, and test set, which contains 1000, 200, and 200 sequences, respectively. The goal is to predict the future 10 steps based on the past 10 observations.

\paragraph{ERA5 Z500} We downsampled the geopotential height at 500 hPa (Z500) from the ERA5 global reanalysis data to a resolution of $2.5^\circ$, resulting in a grid size of $72\times 144$.For our experiments, we utilized data from 2013 to 2019 for training, 2020 as validation, and 2021 and 2022 as test sets, obtaining more than 20,000 sequences. To mitigate the influence of geographic location bias on the Z500 predictions, we implemented a normalization technique by subtracting the Z500 values at each location from the corresponding average from 2013 to 2019. The task is to forecast the subsequent 10 frames of data, predicated on the preceding two observations. This equates to a predictive horizon of 30 hours into the future.

\paragraph{Sea Temperature}

We downloaded 20 years of daily mean sea water potential temperature on the sea surface from the reanalysis ocean data \citeyearpar{oceanreanalysis} provided by ECMWF. For experiments, we use cropped $64\times 64$ temperature data in Atlantic, Indian, and South Pacific for training and validation, to be more exact, from 2000 to 2018 for training with 170,249 sequences and from 2019 to 2020 for validation with 17,758 sequences. Additionally, we use the sea temperature in the North Pacific from 2000 to 2020 for testing, including 65,286 sequences. For each $64\times64$ cropped area, we normalize it in spatial and temporal dimensions to ensure the observations are in a standard distribution, which can make the task free from the noises of sudden change and observation errors, and mainly focus on the dynamics modeling. Since there exists the region shift between training and test sets, this benchmark not only requires the model to capture complex dynamics in the ocean but also maintain good generality. The task is to predict the future 10 frames based on the past 10 observations, corresponding to predicting sea surface temperature in the 10 coming days based on 10 past days' observations.

\paragraph{Spreading Ink} The dataset consists of three open source short videos from \cite{deng2023learning}. The length of three videos are 150, 189, and 139, respectively. Following the experiment setting in Vortex \citeyearpar{deng2023learning}, we split the training and test sets in chronological order by the ratio of 2:1 for each video. Given all the training parts, the goal is to predict all the testing frames at once. For example, for the first video, we need to train our model on the first 100 frames and directly adopt this model to predict all the future 50 frames at once, namely the long-term forecasting task. Since the prediction horizon is much longer than other tasks, this problem poses special challenges in handling accumulative errors. Also, for real fluid video datasets, our concerns also include the model's portrayal of motion continuity and the realism of the generated video.

\subsection{Implementations}\label{appendix:implementation}
In this section, we illustrate the concrete design for incorporating boundary conditions, the aggregation operation, and BFECC with Runge-Kutta Integral within the Multihead Multiscale Integral Architecture.

\paragraph{Boundary Conditions} Here we detail the implementation of incorporating boundary conditions as a supplementary of Eq.~\ref{equ:approximate}. For given boundary conditions $\mathbb{S}$ and the position $\mathbf{r}$, we calculate the correlation on the intersection between boundary $\mathbb{S}$ and $\mathbf{r}$ neighbour $\mathbf{N}_{\mathbf{r}}$. Concretely, we multiply the boundary mask $\mathbbm{1}_{\mathbb{S}}$ to embedded neighbour feature $\widehat{\mathbf{x}}_{T-1}(\mathbf{r}^\prime)$, that is,
\begin{equation}
\begin{aligned}
\label{equ:boundary}
\mathbbm{1}_{\mathbb{S}}(\mathbf{r}^\prime) \left(\widehat{\mathbf{x}}_{T}(\mathbf{r})\cdot\widehat{\mathbf{x}}_{T-1}(\mathbf{r}^\prime)\right) = \left(\widehat{\mathbf{x}}_{T}(\mathbf{r})\cdot\mathbbm{1}_{\mathbb{S}}(\mathbf{r}^\prime)\left(\widehat{\mathbf{x}}_{T-1}(\mathbf{r}^\prime)\right)\right),\ \mathbf{r}^\prime\in\mathbb{V}.
\end{aligned}
\end{equation}
This will preserve the number of neighbor correlation channels, and for $\mathbf{r}^\prime \notin \mathbb{S}$, the correlation values will be set to zero.

\paragraph{Aggregation Operation} Given learned deep representations of prediction $\widehat{\mathbf{x}}_{(T+1)}^l, \widehat{\mathbf{x}}_{(T+1)}^{l+1}$ at the $(l+1)$-th and $l$-th scales, the aggregation operation integrates information between different scales, which can be formalized as follows:
$$ \widehat{\mathbf{x}}_{(T+1)}^{l} = \operatorname{Conv}\left(\operatorname{Concat}\left[\left(\operatorname{Upsample}\left(\widehat{\mathbf{x}}_{(T+1)}^{l+1}\right)\right), \widehat{\mathbf{x}}_{(T+1)}^{l}\right]\right), \  l \ \text{from} \  (L-1) \ \text{to}\  1,$$
where we use bilinear interpolation for the operator $\operatorname{Upsample}(\cdot)$.

\paragraph{BFECC with Runge-Kutta Integral} As mentioned in the main text, the second-order Runge-Kutta method can be expressed as $\operatorname{RK2(\mathbf{r}, \mathbf{v}) = \mathbf{r} + \mathbf{v}(\mathbf{r}+\mathbf{v}(\mathbf{r})\frac{{\rm d}t}{2}){\rm d}t}$, where $\mathbf{r} \in \mathbb{R}^{H\times W}$ and $\mathbf{v} \in \mathbb{R}^{2\times H\times W}$. Relying solely on the Runge-Kutta method for temporal integration may result in error accumulations during advection. To address this issue, BFECC employs a combination of forward and backward integrals, which can be formulated as follows:
\begin{equation}
\begin{aligned}
\label{equ:bfecc}
\mathbf{r}^\prime_{\text{Forth}} &= \operatorname{RK2}(\mathbf{r}, \mathbf{v})\\
\mathbf{r}^\prime_{\text{Back}} &= \operatorname{RK2}(\mathbf{r}^\prime_{\text{Forth}}, -\mathbf{v})\\
\Tilde{\mathbf{r}} &= \mathbf{r} + \frac{\mathbf{r} - \mathbf{r}^\prime_{\text{Back}}}{2}\\
\operatorname{BFECC}(\mathbf{r}, \mathbf{v}) &= \operatorname{RK2}(\Tilde{\mathbf{r}}, \mathbf{v})
\end{aligned}
\end{equation}

For a given point $\mathbf{r}$, we initially integrate with the velocity vector $\mathbf{v}$ to obtain $\mathbf{r}_{\text{Forth}}^\prime$. Subsequently, we utilize $\mathbf{r}_{\text{Forth}}^\prime$ in conjunction with $-\mathbf{v}$ to perform another integral, resulting in $\mathbf{r}^\prime_{\text{Back}}$. The disparity between $\mathbf{r}$ and $\mathbf{r}^\prime_{\text{Back}}$ signifies the deviation between forward and backward integrals, with half of this difference employed to correct the initial position $\mathbf{r}$.

\subsection{Metrics and Standard Deviations}

In all four datasets, we report the mean value of relative L2 of three repeated experiments with different random seeds as a main metric. Experimentally, the standard deviations of relative L2 are smaller than 0.001 for Navier-Stokes, Bounded N-S and Sea temperature and smaller than 0.003 for Spreading Ink. For scientific rigor, we keep four decimal places for all results. For the Sea Temperature dataset, we report the MSE loss following the common practice in meteorological forecasting. For the Spreading Ink dataset, we used VGG Perceptual Loss \citep{johnson2016perceptual} to measure the realism of the generated fluids. Given $n$ step predictions $\{\widehat{\mathbf{x}}_i\}_{i=1,\cdots,n}$ and corresponding ground truth $\{\mathbf{x}_i\}_{i=1,\cdots,n}$, $\widehat{\mathbf{x}}_i,\mathbf{x}_i\in\mathbb{R}^{H\times W}$, the above-mentioned metrics can be calculated as follows: 
\begin{equation*}
\operatorname{MSE} = \frac{1}{n}\sum_{i=1}^n \frac{1}{H \times W} \|\mathbf{x}_i - \widehat{\mathbf{x}}_i\|_2^2, \ \ \operatorname{Relative\ L2\ Loss} = \frac{ \sqrt{\sum_{i=1}^n \|\mathbf{x}_i - \widehat{\mathbf{x}}_i\|_2^2}} {\sqrt{\sum_{i=1}^n \|\mathbf{x}_i\|_2^2}}.
\end{equation*}
Specifically, for the Spreading Ink dataset and Bounded N-S dataset with prescribed boundary conditions $\mathbb{S}$, we exclusively calculate the loss function within the specified boundary and subsequently report the average. Let $\mathbb{D}$ denote the region inside the container, and thus, MSE and Relative L2 can be computed as follows:
\begin{equation*}
\operatorname{MSE} = \frac{1}{n}\sum_{i=1}^n \frac{1}{|\mathbb{D}|}\sum_{(j,k)\in\mathbb{D}}(\mathbf{x}_{ijk} - \widehat{\mathbf{x}}_{ijk})^2, \ \ \operatorname{Relative\ L2\ Loss} = \frac{ \sqrt{\sum_{i=1}^n \sum_{(j,k)\in\mathbb{D}}(\mathbf{x}_{ijk} - \widehat{\mathbf{x}}_{ijk})^2}} {\sqrt{\sum_{i=1}^n \sum_{(j,k)\in\mathbb{D}}\mathbf{x}_{ijk}^2}},
\end{equation*}
where $\mathbf{x}_{ijk}$ represents the value at position $(j,k)$ of $i-$th frame, and $|\mathbb{D}|$ represents the number of grid points in $\mathbb{D}$.

\subsection{Model and Experiment Configurations}

All the experiments are implemented in PyTorch\citep{Paszke2019PyTorchAI}, and conducted on a single NVIDIA A100 40GB GPU. We repeat all the experiments three times with random seeds selected from 0 to 1000 and report the average results. We train the model with Adam optimizer \citep{DBLP:journals/corr/KingmaB14} for all baselines. See Table \ref{tab:exp_config_small} for details.

\begin{table}[h]
    \caption{Experiment configurations in HelmFluid for different benchmarks.}
    \label{tab:exp_config_small}
    \vskip 0.1in
    \centering
 \vspace{-5pt}
    \begin{small}
    \begin{sc}
            \renewcommand{\multirowsetup}{\centering}
            \setlength{\tabcolsep}{16pt}
            \begin{tabular}{l|c|c}
                \toprule
                Benchmark & Learning Rate & Batch Size \\
                \midrule
                    Navier-Stokes & $5\times 10^{-5}$ & 10 \\
                    Bounded N-S & $5\times 10^{-5}$ & 5 \\
                    ERA5 Z500 & $5\times 10^{-5}$ & 5 \\
                    Sea Temperature & $5\times 10^{-5}$ & 10 \\
                    Spreading Ink & $5\times 10^{-5}$ & 5 \\
                \bottomrule
            \end{tabular}
    \end{sc}
    \end{small}
\end{table}

In this section, we provide a detailed overview of the model configurations for HelmFluid. Given that fluid dynamics vary across different resolutions, we augment the number of scales for larger inputs, as outlined in Table \ref{tab:model_config_small}. For the Multihead Multiscale Integral Architecture, we adhere to the conventional design principles of U-Net~\citep{ronneberger2015u}, incorporating downsampling, upsampling, and the aggregation of multiscale features.

\begin{table}[h]
    \caption{Hyperparameter configurations of HelmFluid for different resolutions.}
    \label{tab:model_config_small}
    \vskip 0.1in
    \centering
 \vspace{-5pt}
    \begin{small}
    \begin{sc}
            \renewcommand{\multirowsetup}{\centering}
            \setlength{\tabcolsep}{3pt}
            \begin{tabular}{l|c|c}
                \toprule
                Input Resolutions & Hyperparameters & Values \\
                \midrule
             & Number of scales $L$ & $3$ \\
               $64\times 64$ & Number of heads $M$ & $4$ \\
              & Channels of deep representations $\{d_{\text{model}}^{1},\cdots,d_{\text{model}}^{L}\}$ & $\{64, 128, 128\}$ \\
             & Number of neighbours to calculate spatiotemporal correlations $|\mathbf{N}_{\mathbf{r}}|$ & $81$ \\
                \midrule
                & Number of scales $L$ & $4$ \\
           $128\times 128$ & Number of heads $M$ & $4$ \\
         $256\times 256$ & Channels of deep representations $\{d_{\text{model}}^{1},\cdots,d_{\text{model}}^{L}\}$ & $\{128, 256, 512, 512\}$ \\
          &  Number of neighbours to calculate spatiotemporal correlations $|\mathbf{N}_{\mathbf{r}}|$ & $81$ \\
                \bottomrule
            \end{tabular}
    \end{sc}
    \end{small}
\end{table}

\section{Ablation Study} \label{sec:ablation}
As a supplementary analysis to the main text, we perform detailed ablations in the quantitative aspect to validate the impact of learning HelmDynamics and accounting for boundary conditions.

\subsection{HelmDynamics Block}
\label{paragraph: helmdynamics_ablation}

In this subsection, we will discuss the design of HelmDynamics Block from three perspectives. First, the necessity of learning velocity from HelmDynamics. Second, the effectiveness of potential and stream functions. Third, the usefulness of multilevel modeling, which we mentioned above, enhances the consistency of velocity fields at different scales. We compare the result on the $64 \times 64$ resolution Navier-Stokes dataset and report relative L2, training time, and GPU memory.

\begin{figure*}[htbp]
\vspace{-10pt}
  \begin{center}
    \includegraphics[width=\textwidth]{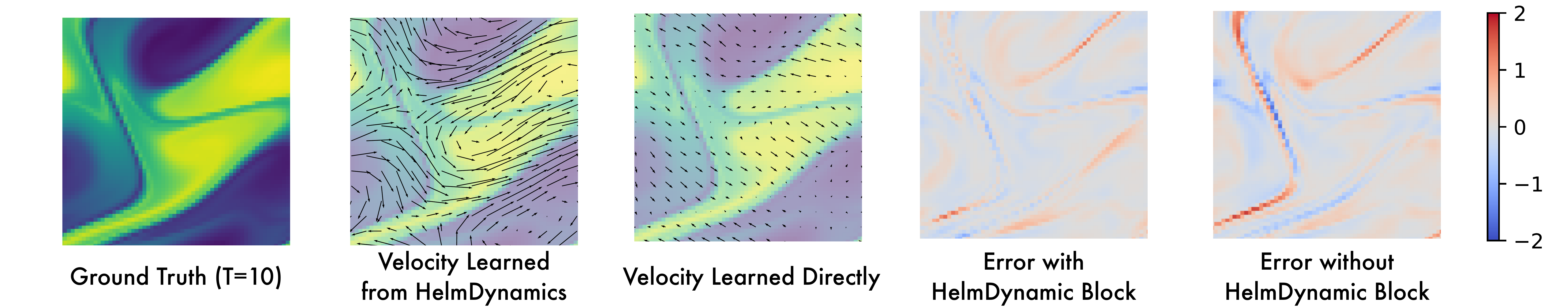}
  \end{center}
  \vspace{-15pt}
  \caption{\small{Velocity field and error comparison between learning by HelmDynamics Block and learning directly.}}\label{fig:ablation_helmdynamics}
  \vspace{-15pt}
\end{figure*}

\paragraph{Learning HelmDynamics or Directly Learning Velocity} We provide the results in Figure \ref{fig:ablation_helmdynamics}, and compare between learning velocity with HelmDynamics and learning velocity directly. We discover that directly learning the superficial velocity will overwhelm the model from capturing complex fluid interactions. As presented in Table \ref{tab:ablation_dynamics}, without Helmholtz dynamics, the performance decreases from 0.1261 to 0.1412, demonstrating the effectiveness of our proposed Helmholtz dynamics. In addition, the calculation of HelmDynamics only brings marginal extra computation costs.

\begin{table*}[h]
        \centering
        \caption{Ablations on dynamics learning in $64\times 64$ Navier-Stokes Dataset.}\label{tab:ablation_dynamics}
        \vspace{5pt}
    \begin{small}
        \begin{sc}
    \begin{threeparttable}
            \renewcommand{\multirowsetup}{\centering}
            \setlength{\tabcolsep}{7pt}
            \begin{tabular}{l|cc|cc}
                \toprule
                    & \multicolumn{2}{c}{Multihead version} & \multicolumn{2}{c}{Single head version} \\
                     \cmidrule(lr){2-3}\cmidrule(lr){4-5}
                Metrics & Velocity & HelmDynamics &  Velocity & HelmDynamics \\
                \midrule
            Relative L2 & 0.1412  &\textbf{0.1261} &  0.1461 & \textbf{0.1344} \\
            \midrule
            GPU memory (GB) & 14.86 & 16.30 & 13.02 & 14.41 \\
            Training Time (s / epoch) & 72.18 & 80.20 & 48.25 & 61.22 \\
                \bottomrule
            \end{tabular}
 \end{threeparttable}
 
        \end{sc}
    \end{small}
\end{table*}

\paragraph{Are Both Potential and Stream Functions Effective?} As presented in Table \ref{tab:ablation_helmcomp_new}, only learning potential function or stream function will cause a decrease in the final performance, demonstrating the effectiveness of both components.

\begin{table*}[h]
        \centering
        \caption{Ablations on learning HelmDynamics, single potential or stream function in 64 $\times$ 64 Navier-Stokes Dataset.}\label{tab:ablation_helmcomp_new}
        
        \vspace{5pt}
    \begin{small}
        \begin{sc}
    \begin{threeparttable}
            \renewcommand{\multirowsetup}{\centering}
            \setlength{\tabcolsep}{7pt}
            \begin{tabular}{l|c|c|c}
                \toprule
                Metrics & HelmDynamics & Only potential function & Only stream function  \\
                \midrule
            Relative L2 & \textbf{0.1261} & 0.1460 & 0.1305 \\
            \midrule
            GPU memory (GB) & 16.30 & 16.29 & 16.30 \\
            Training Time (s / epoch) & 80.20 & 79.57 & 79.60 \\
                \bottomrule
            \end{tabular}
 \end{threeparttable}
 
        \end{sc}
    \end{small}
\end{table*}

\paragraph{Learning HelmDynamics in Multiple Scales}
As presented in Eq.~\ref{equ:process}, we ensemble the learned HelmDynamics in multiple scales. Here we also provide ablations on just employing HelmDynamics in one single scale in Table \ref{tab:ablation_multi}. We can find that our multiscale design can facilitate the dynamics modeling.

\begin{table}[htbp]
        \centering
        \caption{Ablations on learning HelmDynamics in multiple or single scales.}\label{tab:ablation_multi}
        \vspace{5pt}
    \begin{small}
        \begin{sc}
    \begin{threeparttable}
            \renewcommand{\multirowsetup}{\centering}
            \setlength{\tabcolsep}{7pt}
            \begin{tabular}{l|c|c|c}
                \toprule
                Metrics & Multiple Scales & Single Scale (Bottom) & Single Scale (Top) \\
                \midrule
            Relative L2 & \textbf{0.1261} & 0.1441 & 0.1798\\
            \midrule
            GPU memory (GB) & 16.30 & 8.80 & 11.68 \\
            Training Time (s / epoch) & 80.20 & 30.69 & 45.32 \\
                \bottomrule
            \end{tabular}
    \end{threeparttable}
 
        \end{sc}
    \end{small}
    \vspace{-5pt}
\end{table}

\paragraph{Number of neighbors in correlation calculation}
A larger regional area will provide more information for spatiotemporal correlation calculation. In this paper, we choose $|\mathbf{N}_{\mathbf{r}}|$ as $9\times 9$. Here in Table \ref{tab:ablation_neighbours}, we also compared with $7 \times 7$, $5 \times 5$, and $3 \times 3$. The model parameters are quite close for several neighbor sizes, but the best performance is obtained for $9\times 9$.

\begin{table}[htbp]
        \centering
        \caption{Ablations on the number of neighbors in correlation calculation.}\label{tab:ablation_neighbours}
        
        \vspace{5pt}
    \begin{small}
        \begin{sc}
    \begin{threeparttable}
            \renewcommand{\multirowsetup}{\centering}
            \setlength{\tabcolsep}{7pt}
            \begin{tabular}{l|cccc}
                \toprule
                \textbf{Number of neighbours in correlation $|\mathbf{N}_{\mathbf{r}}|$} & 3$\times$3 & 5$\times$5 &7$\times$7 & {9$\times$9}  \\ 
                \midrule
                Parameter number & 9,825,421 & 9,848,461 & 9,883,021 & 9,929,101 \\
                \midrule
                Relative L2 & 0.1337 & 0.1273 & 0.1272 & \textbf{0.1261} \\
                \bottomrule
            \end{tabular}
 \end{threeparttable}
 
        \end{sc}
    \end{small}
        \vspace{-10pt}
\end{table}

\subsection{Boundary Conditions}
\label{subsection: boundaryconditions_ablation}

\begin{figure*}[h]
\vspace{-5pt}
  \begin{center}
    \includegraphics[width=\textwidth]{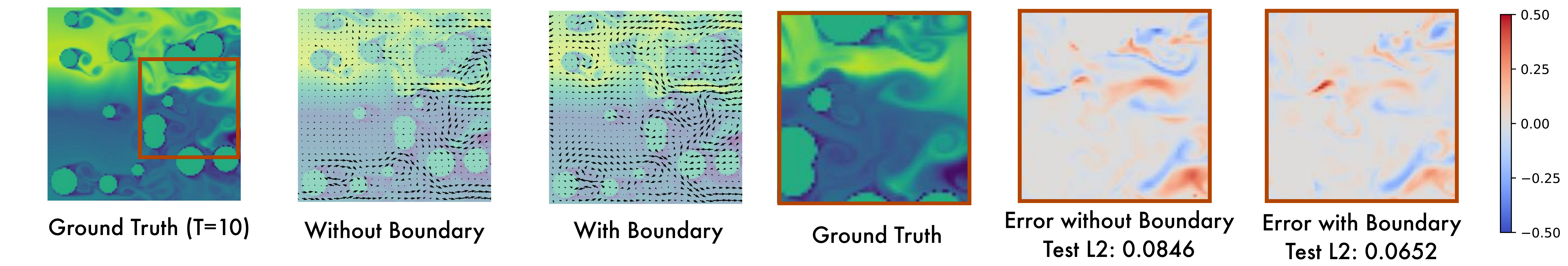}
  \end{center}
  \vspace{-15pt}
  \caption{\small{Velocity field and error comparison between learning by HelmDynamics Block and learning directly.}}\label{fig:ablation_boundary}
  \vspace{-20pt}
\end{figure*}

In this subsection, we discuss the design for boundary condition in correlation calculation, compare the result on the Bounded N-S dataset, and report relative L2, training time and GPU memory. As shown in Figure \ref{fig:ablation_boundary}, while omitting boundary conditions, the learned velocities are perpendicular to the boundary, leading to discontinuous predictions. We present the quantitative results in Table \ref{tab:ablation_boundary}, without input boundary condition, the performance drops seriously, indicating the necessity of our design in HelmDyanmics. It is also notable that as a flexible module, it is quite convenient to incorporate boundary conditions into the HelmDyanmics block, which is also a unique advantage of our model against others.
\begin{table*}[!htbp]
        \centering
        \vspace{-10pt}
        \caption{Ablations on boundary conditions in the Bounded N-S dataset.}\label{tab:ablation_boundary}
        \vspace{5pt}
    \begin{small}
        \begin{sc}
    \begin{threeparttable}
            \renewcommand{\multirowsetup}{\centering}
            \setlength{\tabcolsep}{7pt}
            \begin{tabular}{l|c|c}
                \toprule
                Metrics & Omitting Boundary Conditions & Using Boundary Conditions \\
                \midrule
            Relative L2 & 0.0846 &  \textbf{0.0652} \\
            \midrule
            GPU memory (GB) & 26.98 & 29.48 \\
            Training Time (s / epoch) & 226.20 & 267.63 \\
                \bottomrule
            \end{tabular}
 \end{threeparttable}
        \end{sc}
    \end{small}
        \vspace{-15pt}
\end{table*}

\subsection{Multiscale Multihead Integral Architecture}

We include a summary of ablations for multiscale multihead integral architecture in Table \ref{tab:hyperparam}, including order of runge-kutta for temporal integral, number of heads and number of scales in multiscale multihead integral architecture. We compare the results on the Navier-Stokes dataset of 64 × 64 resolution and report relative L2, training time, and GPU memory.
\paragraph{Order of Runge-Kutta for temporal integral}
Runge-Kutta methods are widely employed for iteratively solving PDEs. The accuracy of the results increases with a higher number of orders but at the cost of additional computation time. In HelmFluid, the accuracy of prediction results is contingent on the precision of the velocity obtained through HelmDynamics blocks. According to our experiments, the second-order Runge-Kutta method is already sufficient for temporal integral, while less order leads to inaccuracies, and more order leads to about $10\%$ more training time. Thus, we choose the second-order Runge-Kutta for integral to trade off performance and efficiency.

\paragraph{Number of heads in multihead modeling}
Adding heads is a convention to augment model capacity \cite{NIPS2017_3f5ee243}, and in fluid prediction, different heads can capture different dynamic patterns. In this paper, adding heads also means more operations in conducting integral. We set $M$ as 4 for a good balance of running time and performance.

\paragraph{Number of scales in multiscale modeling}
This hyperparameter is highly related to the nature of fluid. Considering both model efficiency and fluid dynamics, we choose $L$ as 3 for $64\times 64$ inputs and $4$ for larger inputs.

\begin{table*}[t]
 \caption{Model performances on Navier-Stokes Dataset of $64\times64$ resolution with different selections for orders of Runge-Kutta, number of neighbors in correlations, number of heads and number of scales in multiscale multihead integral architecture.} The \textcolor{MyDarkRed}{red} marked hyperparameter represents the final configuration of HelmFluid.	\label{tab:hyperparam}
    \centering
    \begin{small}
        \begin{sc}
        \vspace{5pt}
            \renewcommand{\multirowsetup}{\centering}
            \setlength{\tabcolsep}{8pt}
            \scalebox{1}{
            \begin{tabular}{l|cccc}
                \toprule
                \textbf{Order of Runge-Kutta} & 1 & \textcolor{MyDarkRed}{2} & 3 & 4 \\
                \midrule
                Relative L2  & 0.1298 & \textbf{0.1261} & 0.1268 & 0.1278  \\
                \midrule
                Training time (s / epoch) & 80.04 & 81.20 & 88.30 & 90.49 \\
                \toprule
                   \textbf{Number of heads $M$} & 1 & \textcolor{MyDarkRed}{4} & 8 & 16  \\
                \midrule
                Parameter number & 11,063,245 & 9,929,101 & 9,812,653 & 9,762,205  \\
                \midrule
                Relative L2  & 0.1344 & \underline{0.1261} & 0.1279 & \textbf{0.1249}  \\
                \midrule
                Training time (s / epoch) & 59.69 & 81.20 & 120.86 & 171.97 \\
                \toprule
                    \textbf{Number of scales $L$}  & 2 & \textcolor{MyDarkRed}{3} & 4 & 5  \\
                \midrule
                Parameter number & 9,283,977 & 9,929,101 & 15,906,193 & 29,820,309  \\
                \midrule
                Relative L2  & 0.1514 & \textbf{0.1261} & 0.1361 & \underline{0.1330}  \\
                \midrule
                Training time (s / epoch) & 64.43 & 81.20 & 99.83 & 120.06 \\
                \bottomrule
            \end{tabular}}
        \end{sc}
    \end{small}
\end{table*}

\subsection{Sensitivity to the Number of Parameters}
We also add the sensitivity analysis to the number of parameters on the $64\times 64$ Navier-Stokes dataset in Table \ref{tab:ablation_para}. We report the results of changing the channels of deep representations to a half and twice the original channels. These results show that the original configuration can achieve a favorable balance between performance of efficiency.
\begin{table*}[h]
        \centering
        \caption{Ablations on the number of parameters.}\label{tab:ablation_para} 
        \vspace{5pt}
    \begin{small}
        \begin{sc}
    \begin{threeparttable}
            \renewcommand{\multirowsetup}{\centering}
            \setlength{\tabcolsep}{5pt}
            \begin{tabular}{l|c|c|c}
                \toprule
                Channels compared to the official configuration & $1/2$ & 1 & 2  \\
                \midrule
            Relative L2  & 0.1380 & \underline{0.1261} & \textbf{0.1242}\\
            \midrule
            GPU memory (GB)  & 9.64 & 16.30 & 29.99 \\
            Running Time (s / epoch) & 75.10 & 80.20 & 112.13\\
            \#Parameter & 2,516,173 & 9,929,101 & 39,446,029\\
                \bottomrule
            \end{tabular}
 \end{threeparttable}
        \end{sc}
    \end{small}
 \vspace{-5pt}
\end{table*}

\section{Additional Results}
\label{sec:additional_results}
\subsection{Efficiency comparision}
\paragraph{Efficiency for different models} In the main text, we presented an efficiency comparison through plots. Here, we provide detailed quantitative results in Table \ref{tab:efficiency} as a supplementary reference.

\begin{table}[h]
        \caption{Efficiency comparison between six deep models on Naiver-Stokes $64\times64$ dataset, where we fixed the batch size to 10.}\label{tab:efficiency}
    \centering
        \vspace{5pt}
    \begin{small}
        \begin{sc}
            \renewcommand{\multirowsetup}{\centering}
            \setlength{\tabcolsep}{5pt}
            \scalebox{1}{
            \begin{tabular}{l|cccccc}
                \toprule
                Models & HelmFluid & U-Net & FNO & MWT & U-NO & LSM \\
                \midrule
                \#Parameter  & 9,929,101 & 17,312,650 & 1,188,641 & 7,989,593 & 61,157,793 & 19,188,033 \\
                \midrule
                Training time (s / epoch) & 80.20 & 46.29 & 18.91 & 90.02 & 103.82 & 44.49 \\
                \midrule
                Relative L2 & \textbf{0.1261} & 0.1982 & 0.1556 & 0.1586 &\underline{0.1435} & 0.1535 \\
                \bottomrule
            \end{tabular}}
        \end{sc}
    \end{small}
 \vspace{-5pt}
\end{table}

\paragraph{Align model size} It's important to note that all baseline models are reproduced using their official configurations from their respective papers, which might lead to an unbalanced model size issue. To ensure a fair comparison, we also scale up the parameters of FNO and compare it with HelmFluid. Refer to Table \ref{tab:align_size} for the results. Notably, even with an increased parameter size comparable to HelmFluid, FNO still exhibits inferior performance.

\begin{table}[h]
\vspace{-10pt}
        \caption{Align model size in 64 $\times$ 64 Navier-Stokes.}\label{tab:align_size}
                \centering
                
        \vspace{5pt}
    \begin{small}
        \begin{sc}
    \begin{threeparttable}
            \renewcommand{\multirowsetup}{\centering}
            \setlength{\tabcolsep}{7pt}
            \begin{tabular}{l|c|c|c}
                \toprule
                Relative L2 & FNO & FNO (enlarged) & HelmFluid \\
                \midrule
            $64\times 64$ Navier-Stokes & 0.1556 & 0.1524 & \textbf{0.1261}\\
            $128\times 128$ Navier-Stokes & 0.1028 & 0.1025 & \textbf{0.0807}\\
            $256\times 256$ Navier-Stokes & 0.1645 & 0.1474 & \textbf{0.1310}\\
                \midrule
            Bounded N-S & 0.1176 & 0.1116 & \textbf{0.0652}\\
                \midrule
            Sea Temperature & 0.1935 & 0.1958 & \textbf{0.1704}\\
                \midrule
            Video 1 & 0.1709 & 0.1872 & \textbf{0.1399}\\
            Video 2 & 0.4864 & 0.5250 & \textbf{0.3565}\\
            Video 3 & 0.1756 & 0.1676 & \textbf{0.1584}\\
            \midrule
            Model Parameter & 1,188,641 & 10,633,265 & 9,929,101  \\
                \bottomrule
            \end{tabular}
 \end{threeparttable} 
        \end{sc}
    \end{small}
\end{table}

\subsection{Full Results for Spreading Ink} We reported the averaged metrics on Spreading Ink dataset. Here, we detail the metrics on three sub-datasets respectively in Table \ref{tab:realresults_full}. Except Relative L2 and MSE are worse than U-Net on Video3, HelmFluid consistently outperforms other models on the other metrics. However, images generated by U-Net appear fragmented, worse than HelmFluid from a visual perspective. Additionally, we present the showcases of three sub-datasets in Figure \ref{fig:real1}, \ref{fig:real2} and \ref{fig:real3}.

\begin{table*}[h]
\vspace{-5pt}
\caption{Model comparison on Spreading Ink for each video. Perceptual loss, Relative L2 and MSE are reported.}
\label{tab:realresults_full}
\vspace{5pt}
\begin{center}
\begin{threeparttable}
\setlength{\tabcolsep}{6.2pt}
\begin{small}
\begin{sc}
\begin{tabular}{l|ccc}
\toprule
Model & Video1  & Video2 & Video3  \\
\midrule
U-Net \citep{ronneberger2015u} & {\underline{1.500} / \underline{0.1544} / \underline{0.0080}} & {\underline{3.982} / 0.4780 / 0.0330} & {\underline{5.307} / \textbf{0.1535} / \textbf{0.0119}} \\
FNO \citep{li2021fourier}&{2.023 / 0.1709 / 0.0097} &{4.732 / 0.4864 / 0.0342} &{5.531 / 0.1756 / 0.0156} \\ 
U-NO \citep{rahman2023uno} & {4.210 / 0.1792 / 0.0106} & {6.204 / 0.5312 / 0.0408} & {6.397 / 0.1810 / 0.0166}\\
Vortex \citep{deng2023learning} & {1.704 / 0.1580 / 0.0083} & {4.171 / \underline{0.4150} / \underline{0.0249}}&{5.973 / 0.1718 / 0.0150} \\
LSM \citep{wu2023solving}  &{1.666 / 0.1592 / 0.0084} &{4.167 / 0.4890 / 0.0346} &{5.448 / 0.1611 / 0.0132} \\ 
\midrule
HelmFluid (Ours) & {\textbf{1.464} / \textbf{0.1399} / \textbf{0.0065}}& {\textbf{3.296} /  \textbf{0.3565} / \textbf{0.0184}}&{\textbf{5.208} / \underline{0.1584} / \underline{0.0127}}  \\
\bottomrule
\end{tabular}
\end{sc}
\end{small}
\end{threeparttable}
\end{center}
\vspace{-10pt}
\end{table*}

\subsection{Align baselines in all benchmarks} As we stated in Section \ref{sec:exp}, some of the baselines are not suitable for part of the benchmarks, specifically Vortex \citep{deng2023learning}, DARTS \cite{ruzanski2011casa}, PWC-Net with fluid Refinement \citep{sun2018pwc} and MWT \citep{Gupta2021MultiwaveletbasedOL}, which means their performance will degenerate seriously or the running time is extremely slow if we stiffly apply them to all benchmarks. Specifically, due to the special design for temporal information in Vortex, we only compare it in the Spreading Ink dataset in the main text. As for the DARTS, since it is designed for the mass field and not applicable for videos with RGB channels, we do not include it in the Spreading Ink dataset. Besides, PWC-Net with fluid Refinement~\citep{zhang2022learning} is proposed to learn the optical flow for fluid, which suffers from the accumulative error, making it far inferior to other methods. Thus, we only compare PWC-Net in the learning velocity field in the main text. 

\begin{figure*}[h]
\centering
    \includegraphics[width=\textwidth]{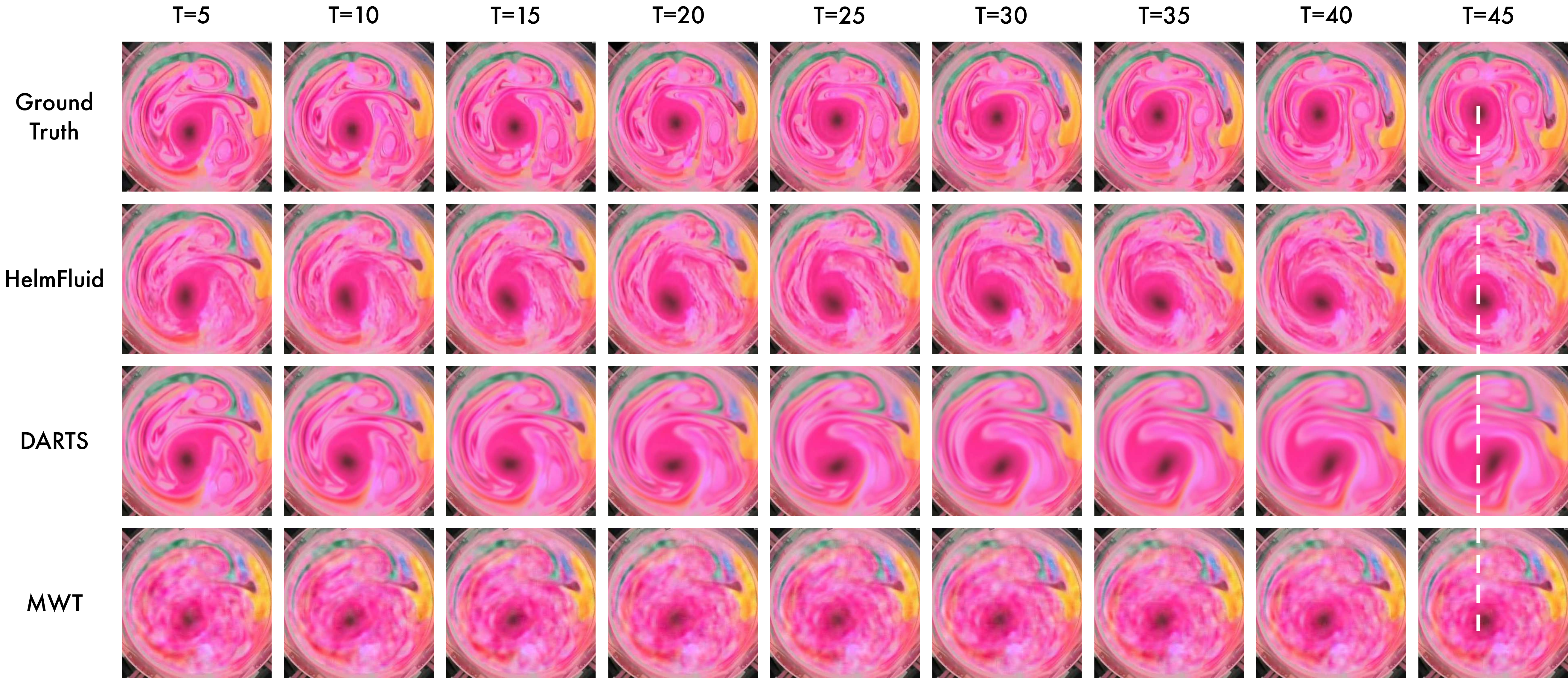}
    \caption{{Showcases of HelmFluid, DARTS, and MWT on the Spread Ink dataset .}}\label{fig:darts_mwt_video3}
\end{figure*}

However, we still provide the missing experiments in Table \ref{tab:align_baseline} to ensure transparency.
\begin{itemize}
    \item Vortex \citep{deng2023learning} models multiple vortex trajectories as a function of time. Since different video sequences have inherently different vortex trajectories, we need to re-train Vortex to fit every video sequence. However, the other three benchmarks except Spreading Ink, have more than 1000 different video sequences. It means that we need to train 1000+ Vortex models for these benchmarks, which is unacceptable. But we still implement this experiment, where we train one vortex model on one single video sequence and generalize it to others.
    \item Due to the slow movement of the spreading ink dataset, DARTS \cite{ruzanski2011casa} showed outstanding quantitative results. However, it fails to predict the correct future in the other three datasets. Moreover, DARTS method solves the least squares problem in the frequency domain for every case, which will bring huge computation costs. In particular, the other deep methods predict the whole sequence in less than 0.1 seconds, while DARTS takes more than 10 seconds. Also, the changes in estimated velocity are very slight with the change of time, which leads to incorrect location estimation. And the extrapolation causes blurring in long-term prediction.
    \item PWC-Net \citep{sun2018pwc} specifically focuses on estimating velocity between adjacent observations. Initially, we attempted to extrapolate predictions using the estimated velocity field, but this approach resulted in severe distortion. To harness the estimated velocity more effectively, we input both the velocity and observations into a U-Net \citep{ronneberger2015u}, yielding improved results denoted as PWC-UNet in Table \ref{tab:align_baseline}. Despite the enhancement provided by the estimated velocity from PWC-Net, U-Net still falls short compared to HelmFluid.
    \item MWT \citep{Gupta2021MultiwaveletbasedOL} predict the future frames based on wavelet analysis. It fails in long-term prediction. The prediction on video 3 of Spreading Ink (Figure \ref{fig:darts_mwt_video3}) shows that as the prediction time gets longer, the prediction image stays at the same position and appears weird texture.
\end{itemize}

\begin{table*}[t]
        \centering
        \caption{Align baselines in all benchmarks, including DARTS \citep{ruzanski2011casa}, adapted version of PWC-Net \citep{sun2018pwc}, MWT \citep{Gupta2021MultiwaveletbasedOL}, Vortex \citep{deng2023learning}. We report Relative L2 for the Navier-Stokes dataset and Bounded N-S dataset, MSE and relative L2 for the Sea Temperature dataset, and Perceptual loss, Relative L2 and MSE for Spreading Ink.}\label{tab:align_baseline} 
        \vspace{5pt}
    \begin{small}
        \begin{sc}
    \begin{threeparttable}
            \renewcommand{\multirowsetup}{\centering}
            \setlength{\tabcolsep}{7pt}
            \begin{tabular}{l|c|c|c|c}
                \toprule
                 & Navier-Stokes & Bounded N-S & Sea Temperature & Spreading Ink (Video 3) \\ 
                \midrule
                DARTS  & 0.8046 & 0.1820 & 0.3308 / 0.1094 & 4.940 /  0.1601 / 0.0130\\
                PWC-UNet  & 0.1765 &  \underline{0.0729} & \underline{0.1805} / \underline{0.0406} & 5.341 / \underline{0.1591} / \underline{0.0128}\\
                MWT  & \underline{0.1586} & 0.1407 & 0.2075 / 0.0510 & \textbf{1.521} / 0.1775 / 0.0160 \\
                Vortex& 8.1379 & 1.6259 & 4.9302 / 0.1796 & 5.973 / 0.1718 / 0.0150\\
                \midrule
                HelmFluid & \textbf{0.1261} & \textbf{0.0652} & \textbf{0.1704} / \textbf{0.0368} &  \underline{5.208} / \textbf{0.1584} / \textbf{0.0127}  \\
            \bottomrule
            \end{tabular}
 \end{threeparttable} 
        \end{sc}
    \end{small}
\end{table*}

\subsection{Performance on turbulence dataset} To effectively show the model performance in handling turbulent fluid, we assess HelmFluid and other baseline models using a turbulence dataset with dimensions of $64 \times 64$~\citep{Wang2019TowardsPD}. This dataset comprises 6000 sequences for training, 1700 for validation, and 2100 for testing. The objective is to predict subsequent velocity fields based on preceding observations. For optimal performance on the dataset, all models are trained with an input sequence of 25 timesteps and evaluated over 20 timesteps. The results are presented in Table \ref{tab:turbulence_dataset}.

\begin{table}[!htbp]
\vspace{-1pt}
        \caption{Performance on turbulence dataset.}\label{tab:turbulence_dataset}
                \centering
                
        \vspace{5pt}
    \begin{small}
        \begin{sc}
    \begin{threeparttable}
            \renewcommand{\multirowsetup}{\centering}
            \setlength{\tabcolsep}{7pt}
            \begin{tabular}{l|c}
                \toprule
                Turbulence dataset & MSE \\
                \midrule
            U-Net \citep{ronneberger2015u}& 1062.13\\
            TF-Net \citep{Wang2019TowardsPD} & \underline{1061.78}\\
            FNO \citep{li2021fourier} & 1187.44\\
            U-NO \citep{rahman2023uno} & 3276.09\\
            LSM \citep{wu2023solving} & 1069.26\\
                \midrule
            HelmFluid & \textbf{1042.38}  \\
                \bottomrule
            \end{tabular}
 \end{threeparttable}
        \end{sc}
    \end{small}
\end{table}

\section{More Showcases}
\label{section:showcases}

As a supplement to the main text, we provide more showcases here for comparison (Figure \ref{fig:NS64}-\ref{fig:real3}). Videos are provided in \underline{Supplementary Materials}.

\begin{figure*}[!htbp]
\centering
    \includegraphics[width=\textwidth]{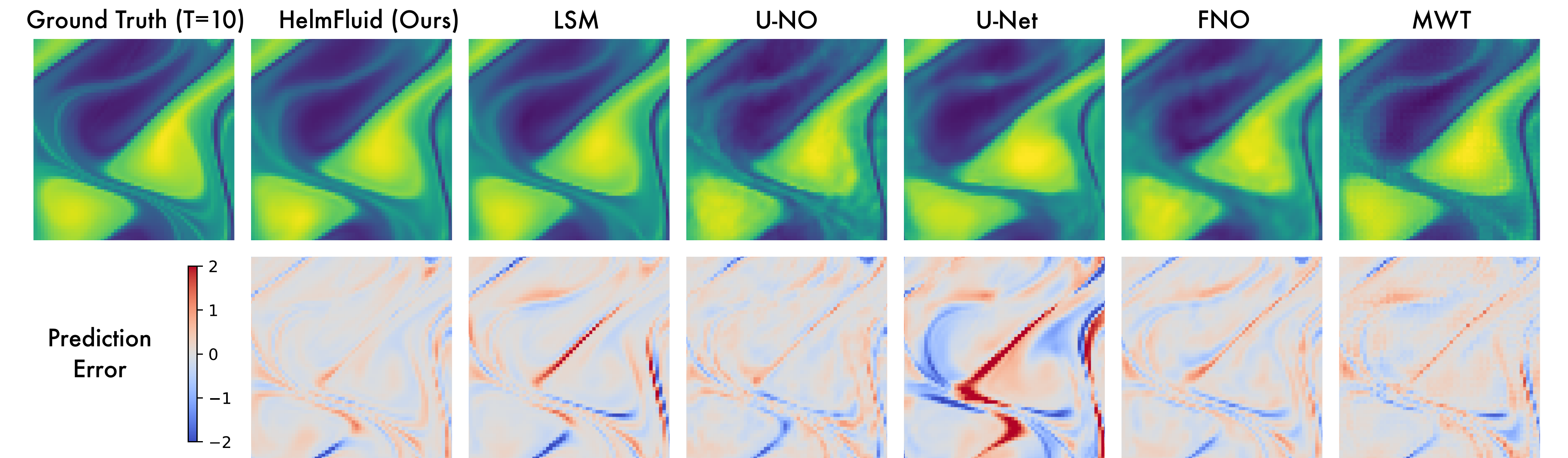}
    \caption{{Showcases of the Navier-Stokes dataset with resolution of $64\times64$.}}\label{fig:NS64}
\end{figure*}

\begin{figure*}[!htbp]
\centering
    \includegraphics[width=\textwidth]{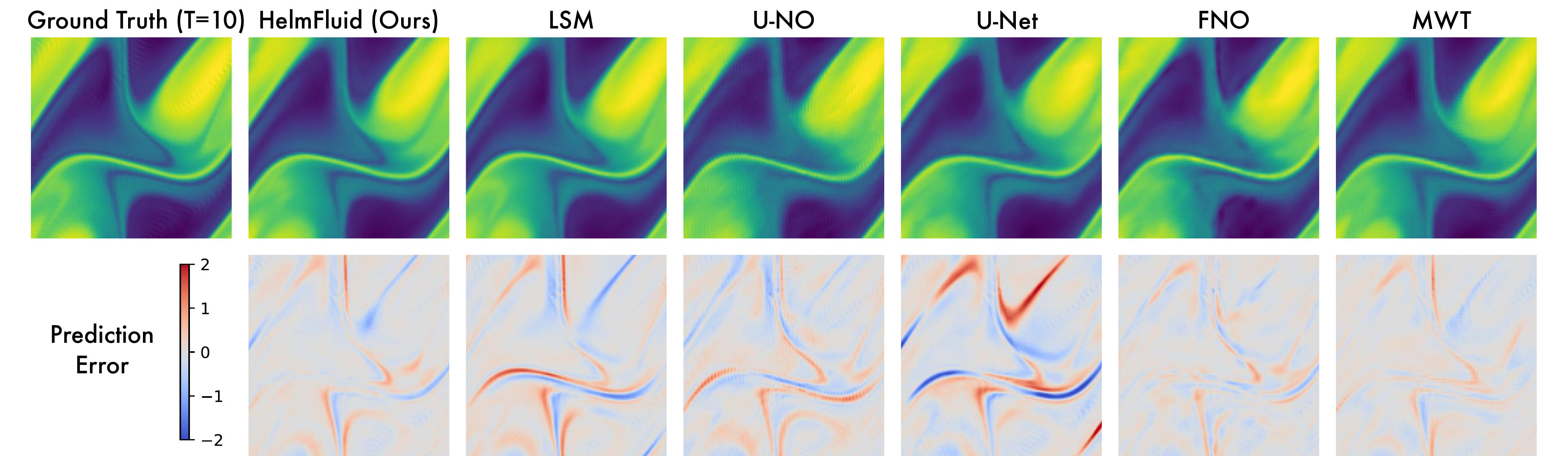}
    \caption{{Showcases of the Navier-Stokes dataset with resolution of $128\times128$.}}\label{fig:NS128}
\end{figure*}

\begin{figure*}[!htbp]
\centering
    \includegraphics[width=\textwidth]{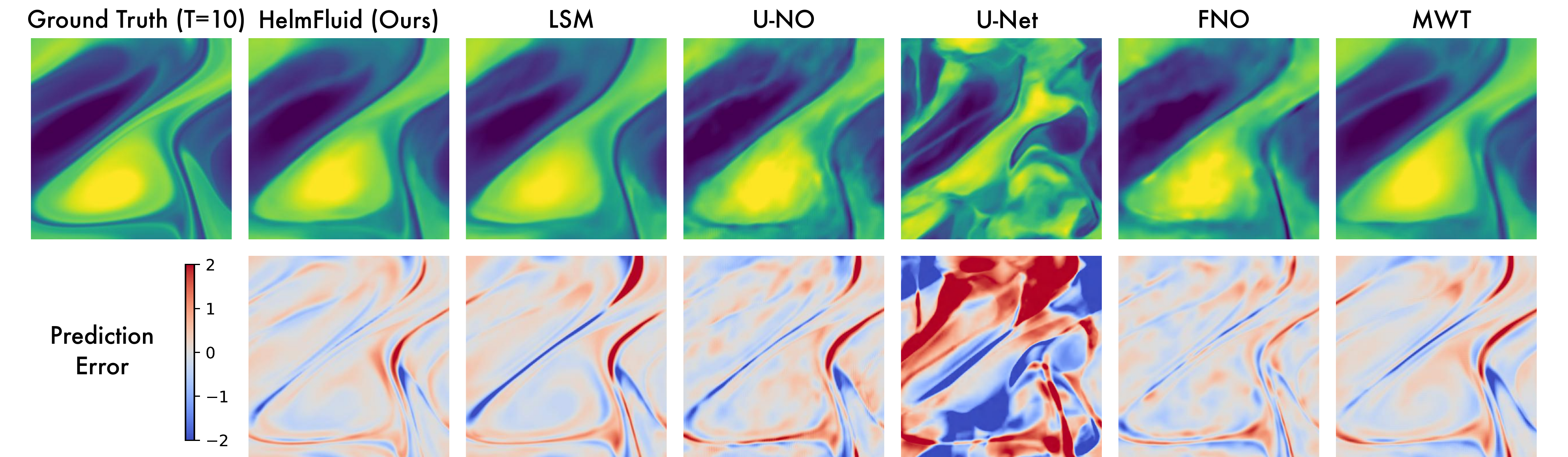}
    \caption{{Showcases of the Navier-Stokes dataset with resolution of $256\times256$.}}\label{fig:NS256}
\end{figure*}

\begin{figure*}[h]
\centering
    \includegraphics[width=\textwidth]{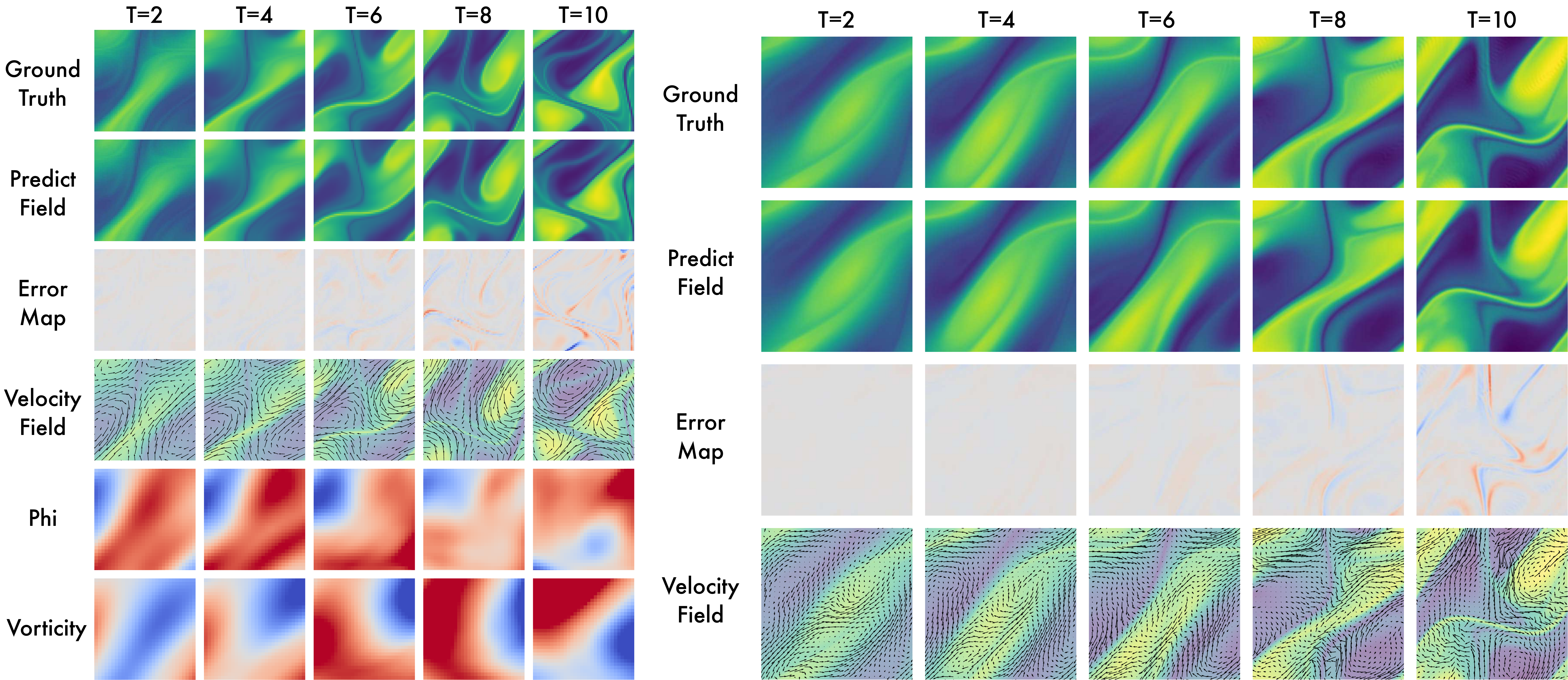}
    \caption{{Showcases of HelmFluid on Navier-Stokes dataset with resolution $64\times64$ and $128\times128$.}}\label{fig:ns_velocity_case}
\end{figure*}

\begin{figure*}[h]
\centering
    \includegraphics[width=\textwidth]{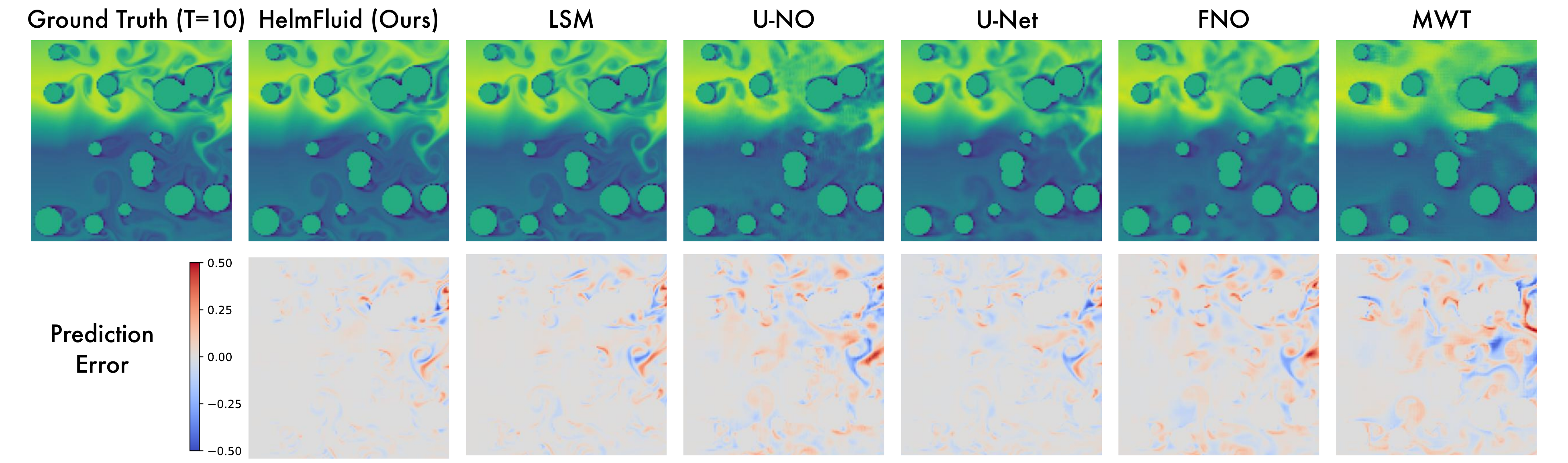}
    \caption{{Showcases of the Bounded N-S dataset.}}\label{fig:Boundedns}
\end{figure*}

\begin{figure*}[h]
\centering
    \includegraphics[width=\textwidth]{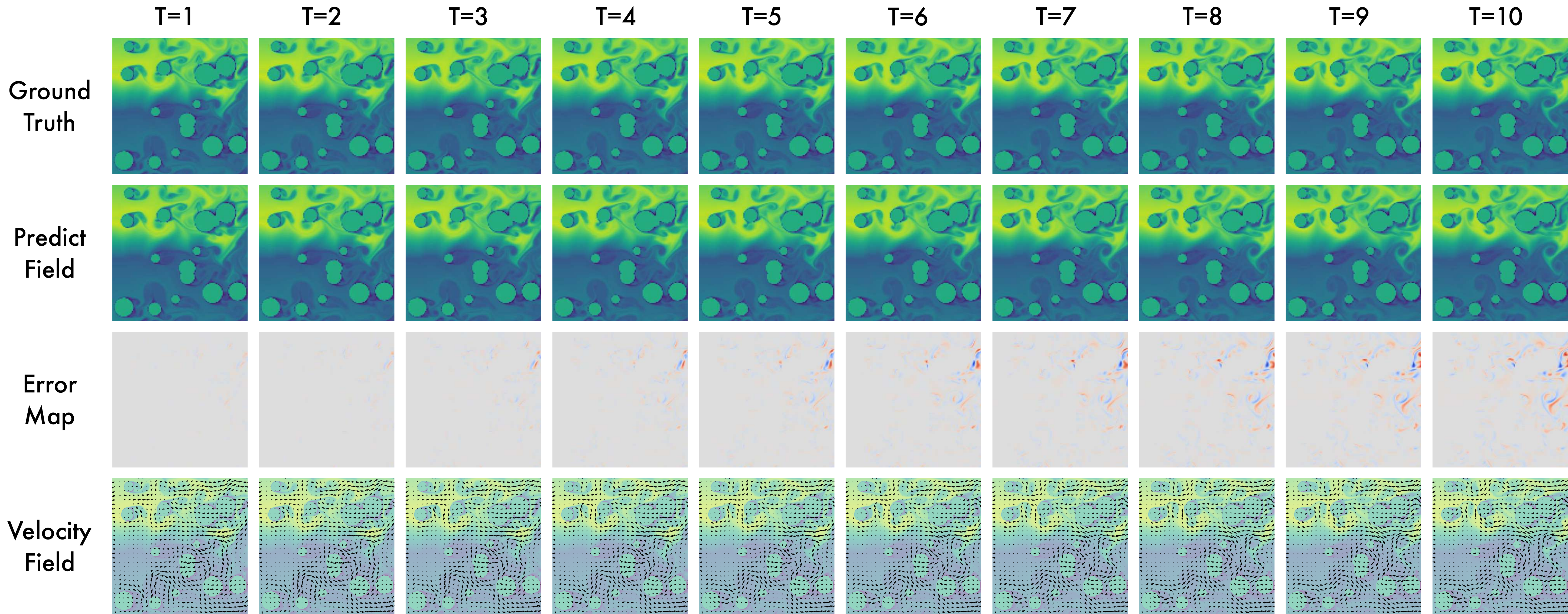}
    \caption{{Showcases of HelmFluid on the Bounded N-S dataset.}}\label{fig:HelmDy_boundedns}
\end{figure*}

\begin{figure*}[h]
\centering
    \includegraphics[width=\textwidth]{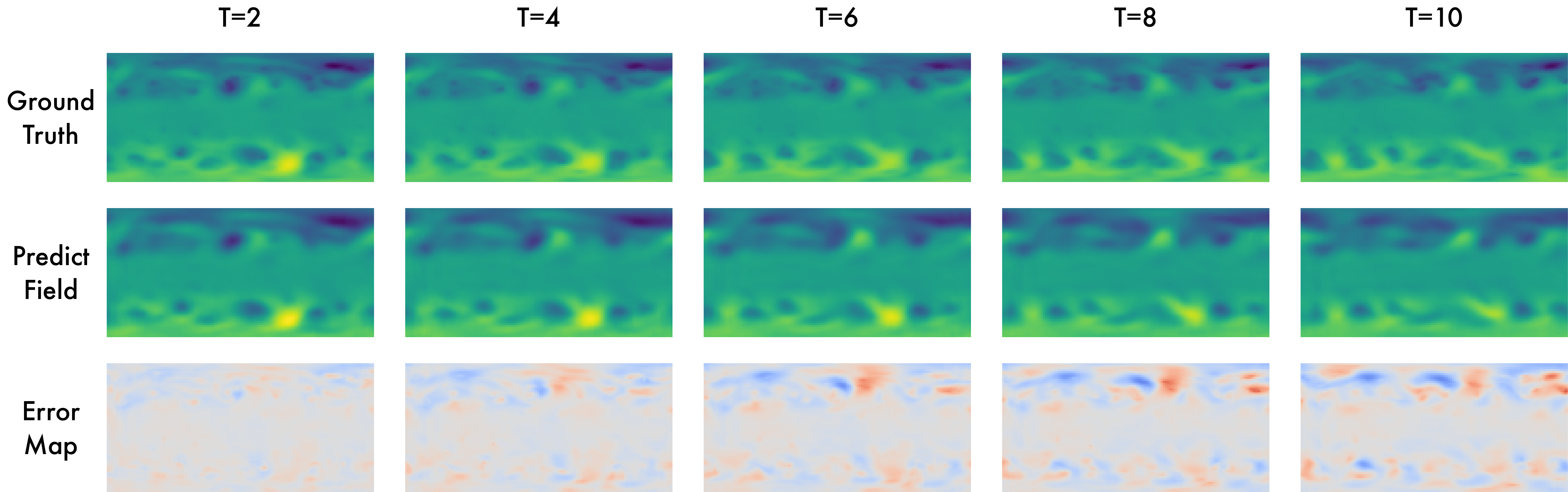}
    \caption{{Showcases of the ERA5 Z500 dataset.}}\label{fig:z500showcase}
\end{figure*}

\begin{figure*}[h]
\centering
    \includegraphics[width=\textwidth]{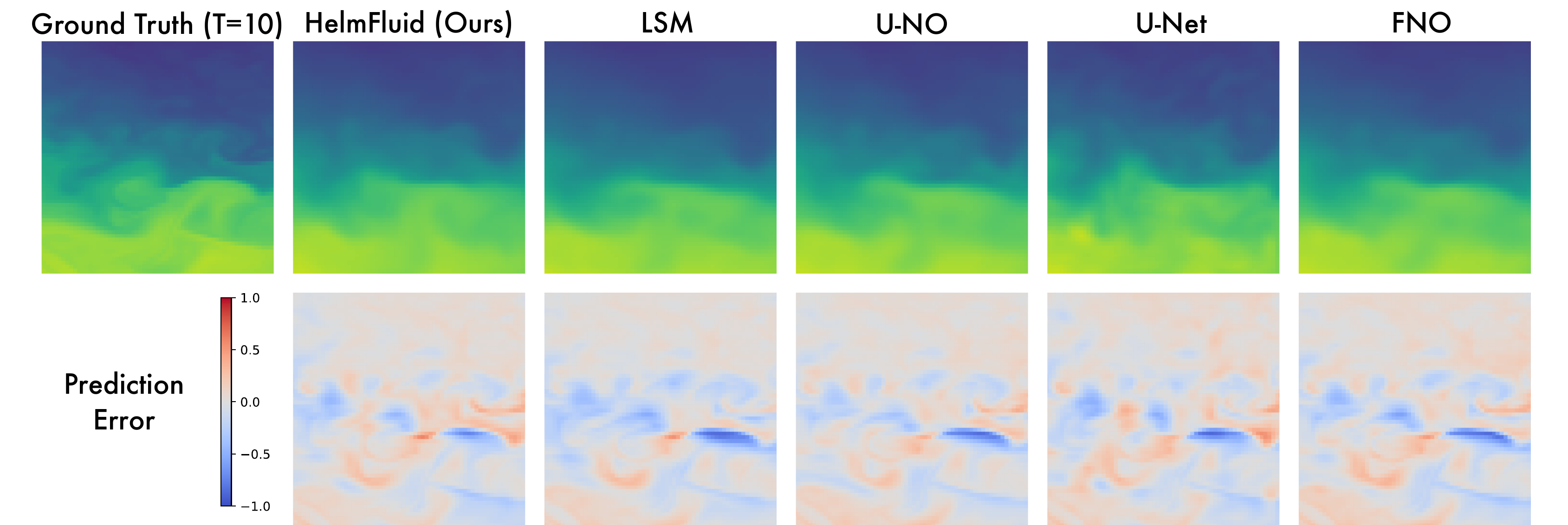}
    \caption{{Showcases of the Sea Temperature dataset.}}\label{fig:sea1}
\end{figure*}

\begin{figure*}[h]
\centering
    \includegraphics[width=\textwidth]{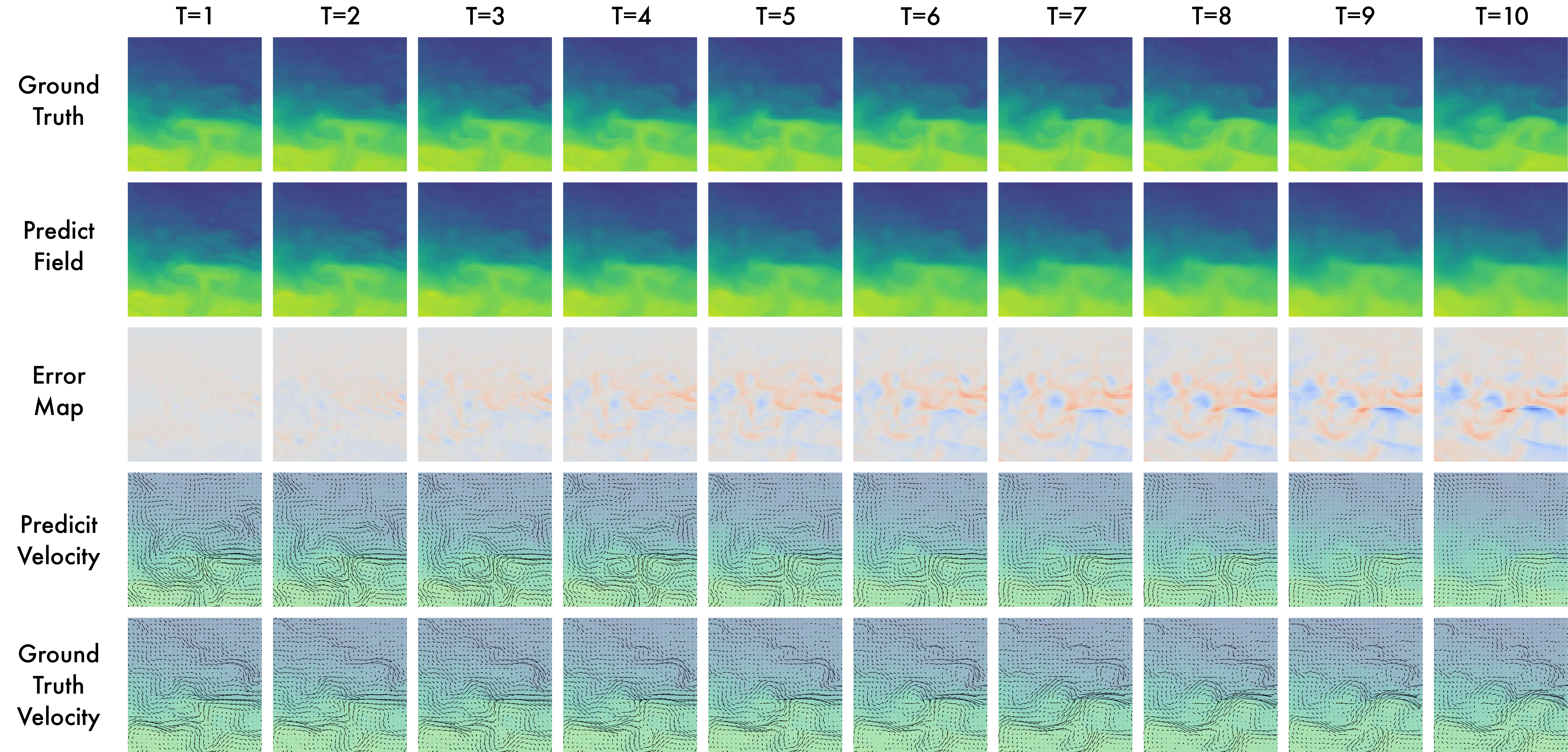}
    \caption{{Showcases of HelmFluid on the Sea Temperature dataset.}}\label{fig:sea2}
\end{figure*}

\begin{figure*}[h]
\centering
    \includegraphics[width=\textwidth]{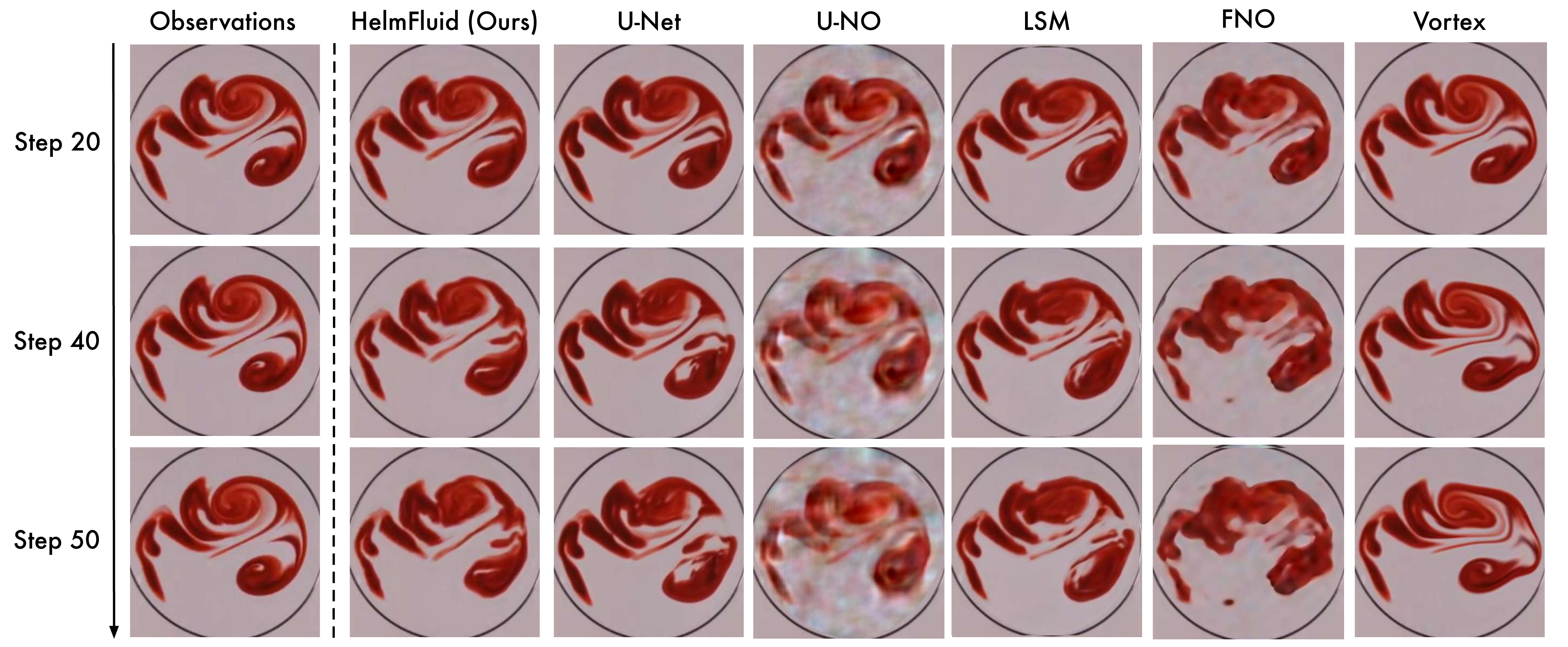}
    \caption{{Showcases of the Spreading Ink dataset (Video 1).}}\label{fig:real1}
\end{figure*}

\begin{figure*}[h]
\centering
    \includegraphics[width=\textwidth]{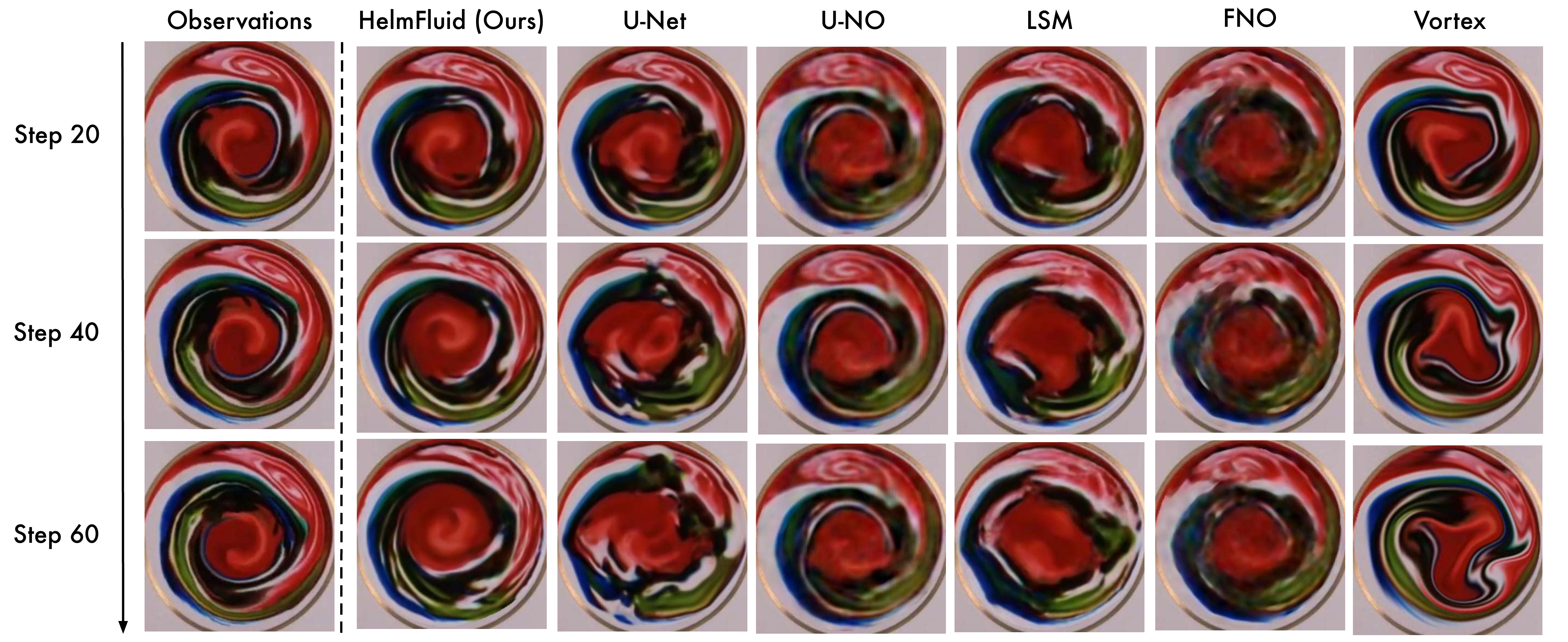}
    \caption{{Showcases of the Spreading Ink dataset (Video 2).}}\label{fig:real2}
\end{figure*}

\begin{figure*}[h]
\centering
    \includegraphics[width=\textwidth]{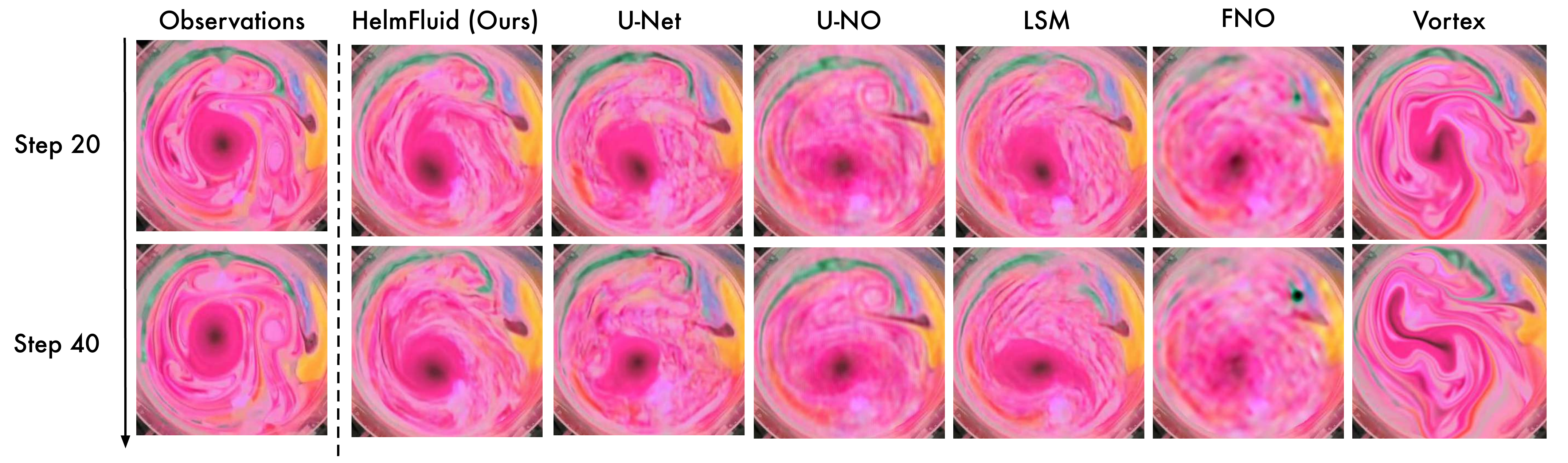}
    \caption{{Showcases of the Spreading Ink dataset (Video 3).}}\label{fig:real3}
\end{figure*}

\end{document}